\documentclass[journal]{IEEEtran}
\usepackage{amsfonts}
\usepackage{bm}
\usepackage{amsmath,amsfonts,amsthm,amssymb}
\usepackage{setspace}
\usepackage{Tabbing}
\usepackage{fancyhdr}
\usepackage{lastpage}
\usepackage{extramarks}
\usepackage{chngpage}
\usepackage{soul,color}
\usepackage{epsfig}
\usepackage{graphicx,float,wrapfig,subfigure}
\usepackage{dsfont}
\usepackage{longtable}
\usepackage{wrapfig}
\usepackage{stfloats}
\usepackage{cite}
\usepackage{graphicx}
\usepackage{array}
\usepackage{multirow}{\tiny }
\usepackage{multicol}
\usepackage{colortbl}
\usepackage{tabularx}
\usepackage{mdwmath}
\usepackage{mdwtab}
\usepackage{color}
\usepackage{verbatim}
\usepackage{amsmath}
\usepackage{tikz}
\usetikzlibrary{calc}
\usepackage{flowchart}
\usetikzlibrary{shapes.geometric, arrows}
\usepackage{overpic}
\usepackage{epstopdf}
\usepackage{url}
\usepackage[colorlinks,linkcolor=black,anchorcolor=black,citecolor=black,urlcolor=black]{hyperref} 

\usepackage{enumitem}
\usepackage[ruled,linesnumbered]{algorithm2e}
\usepackage{makecell}
\usepackage{diagbox}
\usepackage{soul}

\usepackage{color}
\makeatletter %
\let\myorg@bibitem\bibitem
\def\bibitem#1#2\par{%
	\@ifundefined{bibitem@#1}{%
		\myorg@bibitem{#1}#2\par
	}{%
		\begingroup
		\color{\csname bibitem@#1\endcsname}%
		\myorg@bibitem{#1}#2\par
		\endgroup
	}%
}

\makeatother

\allowdisplaybreaks

\usepackage{footnote}
\makesavenoteenv{tabular}
\makesavenoteenv{table}
\SetKwInOut{Input}{Input}\SetKwInOut{Output}{Output}

\begin{document}

\title{
	Customized Load Profiles Synthesis for Electricity Customers Based on Conditional Diffusion Models
	}

\author{
Zhenyi Wang, \IEEEmembership{Graduate Student Member, IEEE,}
and Hongcai Zhang, \IEEEmembership{Senior Member, IEEE}
    \thanks{
    This paper is funded in part by the Science and Technology Development Fund, Macau SAR (File no. 001/2024/SKL, and File no. 0011/2021/AGJ). (Corresponding author: \textit{Hongcai Zhang}.)
    
	Z. Wang, and H. Zhang are with the State Key Laboratory of Internet of Things for Smart City and Department of Electrical and Computer Engineering, University of Macau, Macao, 999078 China (email: hczhang@um.edu.mo).
	}

}

\maketitle

\begin{abstract}
Customers' load profiles are critical resources to support data analytics applications in modern power systems. However, there are usually insufficient historical load profiles for data analysis, due to the collection cost and data privacy issues. 
To address such data shortage problems, load profiles synthesis is an effective technique that provides synthetic training data for customers to build high-performance data-driven models. Nonetheless, it is still challenging to synthesize high-quality load profiles for each customer using generation models trained by the respective customer's data owing to the high heterogeneity of customer load. 
In this paper, we propose a novel customized load profiles synthesis method based on conditional diffusion models for heterogeneous customers. 
Specifically, we first convert the customized synthesis into a conditional data generation issue. 
We then extend traditional diffusion models to conditional diffusion models to realize conditional data generation, which can synthesize exclusive load profiles for each customer according to the customer's load characteristics and application demands.
In addition, to implement conditional diffusion models, we design a noise estimation model with stacked residual layers, which improves the generation performance by using skip connections. The attention mechanism is also utilized to better extract the complex temporal dependency of load profiles.
Finally, numerical case studies based on a public dataset are conducted to validate the effectiveness and superiority of the proposed method.
\end{abstract}

\begin{IEEEkeywords}
Conditional diffusion models, data analytics applications, deep generative model, load profiles synthesis
\end{IEEEkeywords}

\section{Introduction}\label{sec_introduction}
\IEEEPARstart{I}{n} order to build a new generation of power systems to achieve carbon neutrality, it is necessary to improve the digitalization and intelligence of the grid \cite{xie2021toward_carbon}. As an advanced type of metering infrastructure, smart meters are widely deployed to facilitate the integration of energy flow and information flow, consequently improving the efficiency and reliability of power systems \cite{alahakoon2015smart_meter}. On the one hand, relying on a large amount of fine-grained load profiles collected by smart meters, utilities can provide personalized and high-quality services to customers, and reduce their operation costs. On the other hand, customers can be more aware of their power consumption characteristics, save energy costs, and even promote participation in demand response\cite{wang2018review_smart_meter_app, chen2023accurate_cx,wang2024customer}.

However, building data-driven models for data analytics applications, especially deep learning models, usually requires massive amounts of training data\cite{yu2023district_ypp}. Unfortunately, because of the cost, regulatory, and privacy concerns, it is often challenging for utilities, let alone customers, to acquire sufficient training data \cite{lee2020federated_data_shortage}. To tackle data shortages, some researchers have proposed federated learning to collaboratively train models among multiple customers, which has been extensively used in power systems \cite{khan2021federated_survey}. Federated learning can improve the performance of data-driven models, but it requires cooperation among customers to complete tasks. Because customers may have different goals and needs, they are likely to be unwilling to collaborate on certain tasks, which in turn will lead to federated learning not meeting each customer's demands.

In contrast to federated learning, load profiles synthesis does not directly train specialized models for customers. Instead, it generates massive high-fidelity load data for customers so that they can train their own data-driven models. Thus, load profiles synthesis is another effective approach to tackle data deficiency, and many studies have been devoted to this field in recent years. The published methods can generally be divided into two categories:
\textit{model-based} methods and \textit{data-driven} methods. The model-based methods generate load profiles based on mathematical physical modeling of customer loads. López et al. \cite{lopez2018smart_model_physical} simulated residential energy consumption by modeling the physical properties of household appliances and buildings. Diao et al. \cite{diao2017modeling_physical_2} developed a physical model-based load synthesis method for residential buildings, which identifies and simulates the load pattern of each household component. However, these methods require accurate physical models and domain knowledge, which makes them difficult to popularize across different scenarios. Furthermore, model-based methods often adopt approximation techniques, which may result in large deviations in generated load profiles.

To solve the above problems, data-driven approaches that learn the load characteristics based on historical load profiles and then generate new load data for customers, have been widely adopted. Al-Wakeel et al. \cite{al2017k_data_kmeans} proposed a load estimation method based on the k-means algorithm to generate missing and future values of smart meters. Labeeuw et al. \cite{labeeuw2013residential_data_markov} combined the mixture model clustering and Markov models to synthesize residential loads in a top-down manner. Furthermore, with the rapid development of deep learning, some researchers have proposed load profiles generation methods based on deep generative models, especially adversarial generative networks (GAN). For example, El Kababji et al. \cite{el2020data_GAN} proposed a flexible framework based on GAN to generate synthetic labeled load patterns and usage habits at the appliance level. Song et al. \cite{song2022profilesr_HR_GAN} leveraged a GAN-based model and weather data to recover high-resolution load proﬁles from low-resolution load proﬁles through a two-stage process. Huang et al. \cite{huang2022dpwgan_GAN_privacy} developed a Wasserstein GAN approach with differential privacy to transform a real-world dataset into an anonymous synthetic load dataset to preserve customers' data privacy. However, most existing data-driven methods are unconditional, which means that the characteristics and patterns of the load profiles to be synthesized cannot be specified. Thus, these unconditional generation methods cannot satisfy heterogeneous customers' demands to synthesize highly personalized load profiles\cite{chen2023federated_CGAN}.

To address the above issue, conditional information (e.g., season or typical load) needs to be added to the load profiles synthesis process as target guidance. Recently, some studies have focused on conditional data generation in power systems. Chen et al. \cite{chen2023federated_CGAN} presented a novel load generation model using conditional GAN to produce personalized energy consumption data according to the customer label. Moreover, Chen et al. \cite{chen2018model_CGAN_RES} proposed a conditional GAN-based approach for scenario generation, which generates realistic wind and photovoltaic power proﬁles conditioned on weather events or time information. However, these methods can only use discrete values as conditions, which is not suitable for power systems where most information, e.g., historical loads, are continuous 
 \cite{zhang2019scenario_CFGM}.
To aid in data generation using continuous values as conditional information, Zhang et al. \cite{zhang2019scenario_CFGM} proposed ﬂow-based conditional generative models with reversible transformation architectures to forecast residential load profiles according to the previous day's load observations. However, the generalizability of flow-based models is limited because they need specialized architectures designed case-by-case to construct reversible transforms.

To address the above problems, we exploit diffusion models to perform the load profiles synthesis. Diffusion models add random noises to original data through a diffusion Markov process, and then learn to denoise the data via the reverse Markov process for data generation \cite{ho2020denoising_DDPM}. Hence, load profiles can be generated by sampling a random vector from the prior distribution, followed by denoising through the reverse Markov chain. Unlike flow-based models, diffusion models learn the characteristics of load profiles through a fixed Markov process, so they have higher reusability and generalization. Recently, diffusion models have been broadly applied in data generation tasks such as image, speech, and time series synthesis \cite{yang2022diffusion_survey}. However, diffusion models are rarely used in power systems. Therefore, how to adopt diffusion models for customized load profiles synthesis is a valuable and necessary study.

To close the aforementioned research gaps, we propose a novel framework to provide customized load profiles synthesis for varied electricity customers according to their power consumption habits and individual demands. To the best of our knowledge, we are the first to exploit the conditional diffusion models for load profiles synthesis. Specifically, we first formulate the customized load profiles synthesis as the conditional generative issue that considers the load characteristics and requirements of electricity customers. Then, we extend the traditional diffusion models to the conditional diffusion models, and design a deep learning model based on the attention mechanism and residual connection to realize conditional data generation. 
Compared with the published literature, the main contributions of this paper are summarized as follows:
\begin{enumerate}
    \item We propose a customized load profiles synthesis method based on conditional diffusion models. Our proposed method achieves exclusive, high-quality load profiles synthesis for each customer according to their respective continuous conditional information. Unlike popular GAN-based methods, our proposed method can generate more diverse data samples to avoid missing minority modes of historical load profiles. Furthermore, compared with flow-based models, the proposed method has a fixed learning procedure and structure, which is more flexibly adapted to electricity customers with different load patterns, thus being more suitable for customized synthesis.
    \item We design a noise estimation model based on the attention mechanism and residual layers with skip connections to implement conditional diffusion models. The designed model can accurately predict added noises by using the attention mechanism for temporal feature extraction of data, thus enabling excellent data denoising. Moreover, we adopt skip connections to aggregate the output of each residual layer, which enables obtaining high-quality estimation results by considering the information from different estimation stages of the designed model.
\end{enumerate}

The rest of this paper is organized as follows. Section \ref{sec_problem} describes the customized load profiles synthesis problem. The proposed method is elaborated in Section \ref{sec_methodology}. Section \ref{sec_case} validates the effectiveness of the proposed method and Section \ref{sec_conclusion} presents the conclusion and future work of this paper.

\section{Problem Statement}\label{sec_problem}
In this section, we introduce and formulate the problem of customized load profiles synthesis, and then provide two typical realistic application scenarios.

To implement data analytics applications (e.g., load forecasting and anomaly detection), electricity customers usually employ the data-driven model, which requires large amounts of historical load data for training. However, it is often difficult for a single customer to meet that data volume requirement, so customers need synthetic load data to solve the data insufficiency issue. Relying on data-driven models trained with synthetic data, customers can realize their individual applications and targets, e.g., improving energy efficiency or participating in demand response programs.

In this paper, we broadly divide the load profiles synthesis task into the following two scenarios according to whether the customer has load data to train a model:

1) \textit{Data Generation:}
Some customers may have little or even no historical data, rendering them unable to train their data-driven models. They can apply the load profiles synthesis to generate synthetic load data for model training.

2) \textit{Data Augmentation:}
Some customers may have insufficient historical load data that can only train under-fitting data-driven models. In this circumstance, they can apply the load profiles synthesis to augment existing load data and thereby enhance the model performance.

Because load profiles are highly personalized, the electricity customer's load pattern is different from each other. To achieve high-quality data synthesis for each customer, we consider the customized load profiles synthesis that can provide specific generated load data based on the customer's respective load characteristics and target requirements, e.g., the start date and range of load profiles to be synthesized.

For historical load data that can reveal the load pattern of the corresponding customer, we mark them as $\bm{x}_{\text{load}}$. In addition, we suppose that this paper generates the daily load profiles, and denote the date of the load data to be generated as $d$. Considering that $\bm{x}_{\text{load}}$ and $d$ reflect the load characteristic and target requirement, respectively, we intend to fulfill the customized load profiles synthesis as follows:
\begin{equation}\label{problem}
    \bm{y}_{d} = f_{\bm{\theta}}(\bm{x}_{\text{load}}, d),
\end{equation}
where $\bm{y}_{d}$ denotes the generated load profiles for the target date; and $f_{\bm{\theta}}$ is the deep learning model with parameters $\bm{\theta}$.

\section{Proposed Methodology}\label{sec_methodology}
In this section, we elaborate on the proposed method for the customized load profiles synthesis. We first outline the framework and then explain the key techniques in detail.

\subsection{Customized Load Profiles Synthesis Framework}\label{framework}
The goal of our proposed method is to provide customized load profiles synthesis for electricity customers by taking their personalized needs into account. With the help of the generated data, the customers or utilities can build high-performance data-driven models for their specific tasks.

Because electricity customers have varied task demands and historical load data volumes, it seems more reasonable to build a model for each customer individually. However, this approach is costly and may encounter the data shortage problem. Therefore, we exploit the cross-learning approach \cite{toubeau2022privacy_cross} that establishes a generic load profiles synthesis model by treating each customer as a different sample, to carry out customized load synthesis for every customer. 
To accomplish this, we introduce a server that is acted by the utility company or non-profit organization. In addition, we adopt the client-server architecture for the proposed method, as illustrated in Fig. \ref{fig_framework}. Specifically, there are two types of entities in the framework, described as follows:
\begin{figure}
    \centering
    \includegraphics[width=0.9\linewidth]{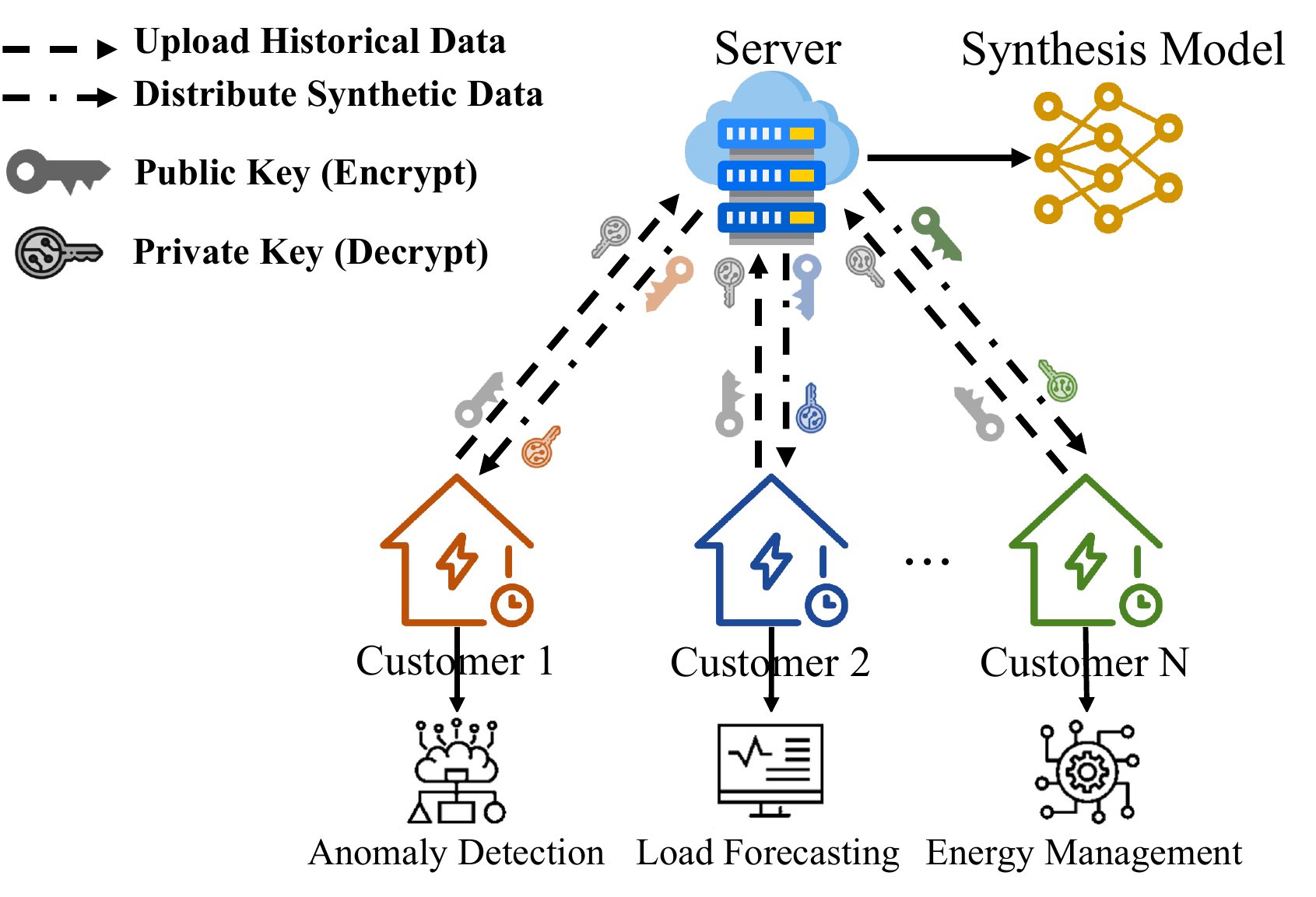}
    \caption{Customized load profiles synthesis framework.}
    \label{fig_framework}
\end{figure}

\subsubsection{Server}
The server gathers load profiles from customers, and trains the global data generation model with the collected data (see details in Section \ref{generation_method}). Moreover, the server provides the customized load profiles synthesis service for each customer based on their personalized requirements, so that they have enough data to build their own data-driven models.

\subsubsection{Customer}
Each customer is responsible for uploading their own historical data to the server to help train the global model. In return, they can obtain customized synthesized load profiles from the server when needed. It is worth noting that uploaded data not only assist in model training but also enable customers to obtain more realistic synthetic load profiles. For new customers with no historical data, the customized data generation service is also available from the server. Although the quality of synthetic data may be slightly poorer, it can preliminarily support their data-driven applications. After attaining the customized synthetic load profiles, customers can build high-performance data-driven models according to their respective application tasks, e.g., load forecasting, anomaly detection, and energy management, etc.


\subsection{Conditional Diffusion Models-Based Generation Method}\label{generation_method}
As the customized synthesis is converted into a conditional data generation form in Eq. (\ref{problem}), we employ diffusion models, which is an advanced deep generative method, to accomplish high-quality load profiles synthesis. The diffusion models are latent variable models that learn the dataset's latent structure to approximate the real data distribution\cite{sohl2015deep_diffusion_models}. What makes diffusion models different from other deep generative methods is that they can generate high-quality samples with diversity, which avoids the instability training problem of GAN-based methods. Moreover, diffusion models have a fixed learning procedure and their latent variable has high dimensionality, so their data generation quality is better and more stable compared to autoencoder-based and flow-based methods\cite{croitoru2022diffusion_comparison}.

\subsubsection{Diffusion Models}
In general, diffusion models are composed of two processes: the \textit{diffusion process} and the \textit{reverse process}. Diffusion models are distinguished from other latent variable models in that the posterior is derived from a Markov chain. Specifically, the Gaussian noise is gradually added to the original data through the diffusion process, which can be formulated as:
\begin{equation}\label{diffusion_process}
    q(\bm{x}_{1:T} \vert \bm{x}_0) := \prod_{t=1}^{T} q(\bm{x}_t \vert \bm{x}_{t-1}),
\end{equation}
where $q(\bm{x}_{1:T} \vert \bm{x}_0)$ is the diffusion process; $\bm{x}_0$ denotes original data sampled from the real data distribution $q(\bm{x}_0)$; $\{\bm{x}_t\}_{t=1}^{T}$ are latent variables (i.e., noisy data) with the same dimensionality as $\bm{x}_0$; $t$ and $T$ are timestep and diffusion total timestep, respectively. Since the diffusion process takes the form of a Markov chain, where $q(\bm{x}_t)$ only depends on $q(\bm{x}_{t-1})$, the joint distribution of noisy data $q(\bm{x}_{1:T})$ can be written as the product of successive diffusion steps. 

For each diffusion step $q(\bm{x}_t \vert \bm{x}_{t-1})$, it takes in $\bm{x}_{t-1}$ and returns $\bm{x}_t$ by adding some noise, which is defined as:
\begin{equation}\label{diffusion_step}
    q(\bm{x}_{t} \vert \bm{x}_{t-1}) := \mathcal{N}(\bm{x}_t; \sqrt{1-\beta_t} \bm{x}_{t-1}, \beta_t \bm{I}),
\end{equation}
where $\beta_{t} \in (0,1)$ is the noise schedule. Finally, $\bm{x}_T$ will follow an isotropic Gaussian distribution when $T \rightarrow \infty$.

Because the diffusion process is deﬁned by a Markov chain, the Gaussian noise is iteratively added to $\bm{x}_0$, so it will cost much time to sample $\bm{x}_t$. To improve the sampling efficiency, we rewrite the diffusion process in Eq. (\ref{diffusion_process}) in a closed form for any arbitrary timestep $t$, using the reparameterization trick:
\begin{equation}\label{diffusion_process_onestep}
    q(\bm{x}_t \vert \bm{x}_0) := \mathcal{N} \big( \bm{x}_t; \sqrt{\overline{\alpha}_t}\, \bm{x}_0, (1-\overline{\alpha}_t)\, \bm{I} \big),
\end{equation}
\begin{equation}\label{diffusion_xt}
    \bm{x}_t := \sqrt{\overline{\alpha}_t}\, \bm{x}_0 + \sqrt{(1-\overline{\alpha}_t)}\, \bm{\epsilon},
\end{equation}
where $\bm{\epsilon} \sim \mathcal{N}(\bm{0}, \bm{I})$ is Gaussian noise; $\overline{\alpha}_t = \prod_{i=1}^{t} \alpha_i$; and $\alpha_t = 1 - \beta_t$.

The reverse process aims to denoise $\bm{x}_t$ to recover $\bm{x}_0$ so that it can eventually recreate data samples from the Gaussian noise. Because the reverse step is hard to directly and explicitly formulate like Eq.(\ref{diffusion_step}), we adopt the neural network model for approximation. Similar to the diffusion step, the reverse step can be written as follows:
\begin{equation}\label{reverse_step}
    p_{\theta} (\bm{x}_{t-1} \vert \bm{x}_t) := \mathcal{N} (\bm{x}_{t-1}; \bm{\mu}_{\theta}(\bm{x}_t, t), \sigma_{\theta} (\bm{x}_t, t) \bm{I}),
\end{equation}
where $p_{\theta} (\bm{x}_{t-1} \vert \bm{x}_t)$ denotes the reverse step implemented by the neural network; $\bm{\mu}_{\theta}(\bm{x}_t, t)$ and $\sigma_{\theta} (\bm{x}_t, t)$ are two neural network models with parameters $\theta$, which take $\bm{x}_t$ and $t$ as inputs and then output the estimated mean and variance of Gaussian distribution, respectively.

Since the reverse process is also deﬁned by a Markov chain, it can be formulated according to the reverse step, as follows:
\begin{equation}\label{reverse_process}
    p_{\theta}(\bm{x}_{0:T}) := p(\bm{x}_T) \prod_{t=1}^{T} p_{\theta} (\bm{x}_{t-1} \vert \bm{x}_t),
\end{equation}
where the joint distribution $p_{\theta}(\bm{x}_{0:T})$ is the reverse process; and $p(\bm{x}_T) = \mathcal{N}(\bm{0}, \bm{I})$ represents the Gaussian noise.

To denoise the data, it is important to accurately estimate the noise added to the original data at each timestep. Given that the reverse process is the inverse of the diffusion process, the objective of the diffusion models can be expressed as:
\begin{equation}\label{loss_naive}
    \min \Big( \text{KL} \big(p_{\theta} (\bm{x}_{t-1} \vert \bm{x}_t), q(\bm{x}_{t-1} \vert \bm{x}_{t}, \bm{x}_0) \big) \Big), 1 \leq t \leq T,
\end{equation}
where $q(\bm{x}_{t-1} \vert \bm{x}_{t}, \bm{x}_0)$ is the posterior conditional probability of the diffusion process (i.e., the ground truth of added noises); and $\text{KL}$ is the abbreviation for Kullback–Leibler divergence, which is to be minimized by using diffusion models.

\subsubsection{Conditional Diffusion Models}
\begin{figure}
    \centering
    \includegraphics[width=0.95\linewidth]{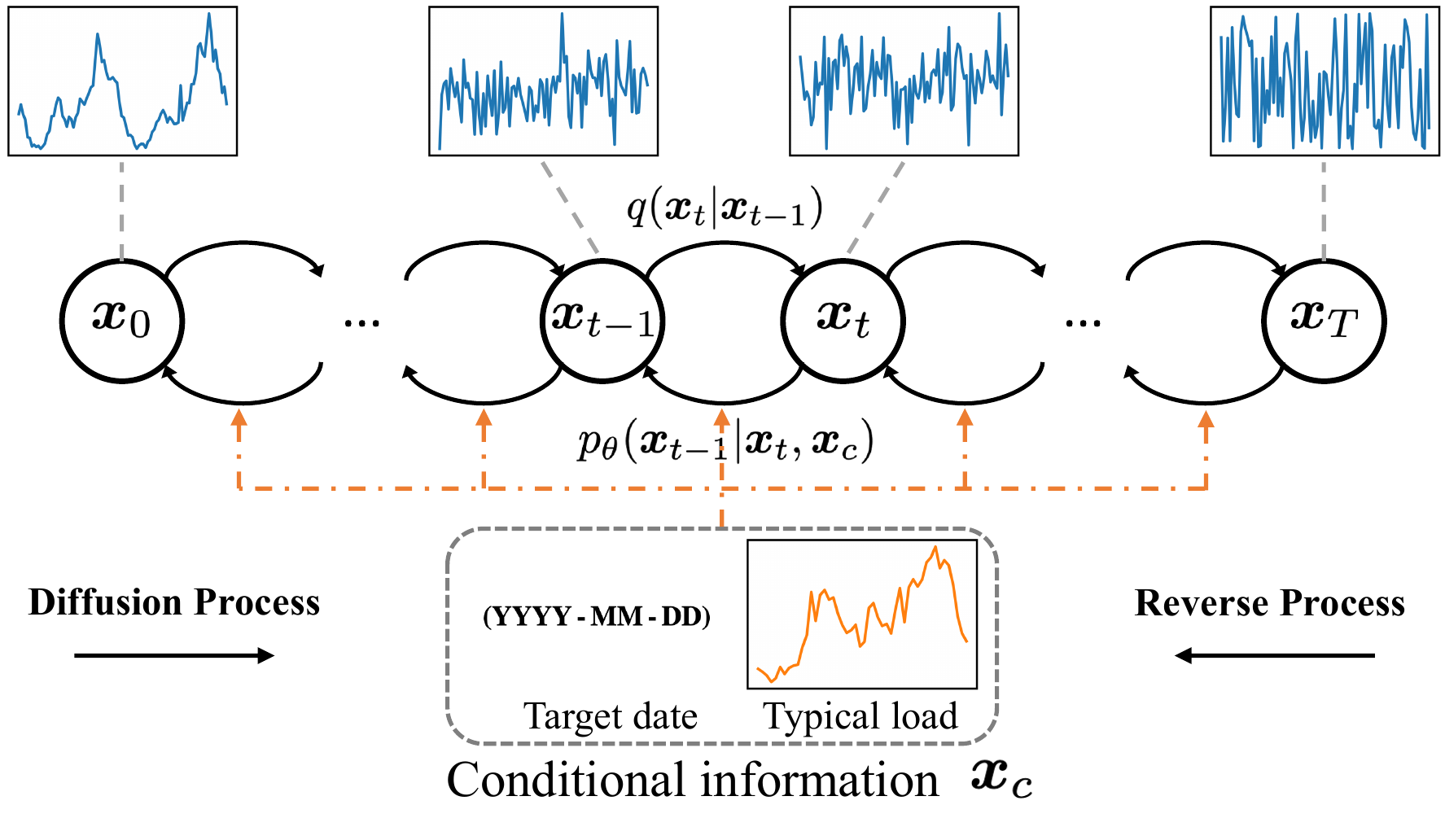}
    \vspace{-4mm}
    \caption{Illustration of conditional diffusion models for customized load profiles synthesis.}
    \label{fig_diffusion}
\end{figure}

To realize the customized synthesis, the proposed method needs to generate specified load data according to each customer's own information and requirements. Therefore, we expand the traditional diffusion models to the conditional one, which is illustrated in Fig. \ref{fig_diffusion}. The conditional diffusion models can synthesize customized load profiles by the guidance of the typical load and target date that implicitly represent the load pattern and individual demand of customers, respectively.

According to Eq. (\ref{reverse_process}), the reverse process of diffusion models can only generate non-specific load profiles because it starts from a Gaussian noise and does not involve any additional information to control data generation \cite{tashiro2021csdi_conditional_DM}. Therefore, we propose conditional diffusion models to consider the customer's personalized demands (i.e., continuous values), which can produce customized load profiles. Specifically, we extend the reverse process to the conditional form, as follows:
\begin{equation}\label{reverse_precess_conditional}
    p_{\theta}(\bm{x}_{0:T} \vert \bm{x}_{\text{c}}) := p(\bm{x}_T) \prod_{t=1}^{T} p_{\theta} (\bm{x}_{t-1} \vert \bm{x}_t, \bm{x}_{\text{c}}),
\end{equation}
\begin{equation}\label{reverse_step_conditional}
    p_{\theta} (\bm{x}_{t-1} \vert \bm{x}_t, \bm{x}_{\text{c}}) := \mathcal{N} (\bm{x}_{t-1}; \bm{\mu}_{\theta}(\bm{x}_t, t \vert \bm{x}_{\text{c}}), \sigma_{\theta} (\bm{x}_t, t \vert \bm{x}_{\text{c}}) \bm{I}),
\end{equation}
where $\bm{x}_c$ denotes the conditional information, which changes with customers, thus accounting for each electricity customer.

Considering that the posterior conditional probability in Eq. (\ref{loss_naive}) represents the noise added to the data, it should follow a Gaussian distribution. To obtain an explicit expression for easy calculation, we rewrite $q(\bm{x}_{t-1} \vert \bm{x}_{t}, \bm{x}_0)$ into a Gaussian form based on Bayes' rule:
\begin{equation}\label{bayes_posterior_conditional}
    q(\bm{x}_{t-1} \vert \bm{x}_{t}, \bm{x}_0) := \mathcal{N}(\bm{x}_{t-1}; \widetilde{\bm{\mu}}_t(\bm{x}_t, \bm{x}_0), \widetilde{\beta}_t\bm{I}),
\end{equation}
\begin{equation}\label{parameter_mu}
    \begin{split}
        \widetilde{\bm{\mu}}_t(\bm{x}_t, \bm{x}_0) &:= \frac{1}{\sqrt{\alpha}_t}(\bm{x}_t - \frac{\beta_t}{\sqrt{1-\overline{\alpha}_t}}\bm{\epsilon}), \\
        \widetilde{\beta}_t &:= \frac{1-\overline{\alpha}_{t-1}}{1-\overline{\alpha}_t} \beta_t.
    \end{split}
\end{equation}
Note that $\widetilde{\bm{\mu}}_t(\bm{x}_t, \bm{x}_0)$ is only related to $\bm{x}_t$ and $\bm{\epsilon}_t$, while $\widetilde{\beta}_t$ is the constant associated with $\beta_t$. We will not discuss $\widetilde{\beta}_t$ subsequently since no unknown variables are involved.

Inspired by the Eq. (\ref{parameter_mu}), the mean expression in the conditional reverse process can be reformulated in a similar way:
\begin{equation}\label{parameter_conditional}
    \bm{\mu}_{\theta}(\bm{x}_t, t \vert \bm{x}_{\text{c}}) := \frac{1}{\sqrt{\alpha_t}} \Big(\bm{x}_t - \frac{\beta_t}{\sqrt{1-\overline{\alpha}_t}} \bm{\epsilon}_{\theta}(\bm{x}_t, t \vert \bm{x}_{\text{c}})\Big),
\end{equation}
where $\bm{\epsilon}_{\theta}$ is the noise estimation model with parameter $\theta$.

Because the KL divergence in Eq. (\ref{loss_naive}) is difficult to calculate directly, we need to convert it into a convenient expression for gradient calculation, which enables the neural network model to update its parameters. Since both $p_{\theta} (\bm{x}_{t-1} \vert \bm{x}_t)$ and $q(\bm{x}_{t-1} \vert \bm{x}_{t}, \bm{x}_0)$ follow a Gaussian distribution, the KL divergence can be calculated in a Rao-Blackwellized fashion with closed form expressions. Therefore, we combine Eqs. (\ref{loss_naive})--(\ref{parameter_conditional}) to formulate the loss function of diffusion models, and we obtain the explicit expressions under the conditional form:
\begin{equation}\label{loss_conditional}
    L(\theta) := \mathbb{E}_{\bm{x}_0 \sim q(\bm{x}_0), \bm{\epsilon} \sim \mathcal{N}(\bm{0}, \bm{I}), t} \ \lVert (\bm{\epsilon} - \bm{\epsilon}_{\theta}(\bm{x}_t, t \vert \bm{x}_{\text{c}})) \rVert_{2}^{2}.
\end{equation}

It is worth noting that we have removed the coefficient of loss function to improve the training quality and stability\cite{ho2020denoising_DDPM}.

\textbf{Remark 1.} \textit{The customized load profiles synthesis is formulated as a conditional data generation problem and intended to be solved by the proposed conditional diffusion models. Based on the customer's conditional information, the well-trained neural network model of the conditional reverse process can denoise from Gaussian noise to generate load profiles specific to that customer, which enables the high-quality customized synthesis. To achieve excellent denoising, the challenge now is to accurately predict the noise added during the diffusion process, which will be addressed in Section \ref{noise_model}.}

\subsection{Attention Mechanism-Based Noise Estimation Model}\label{noise_model}
\begin{figure}
    \centering
    \includegraphics[width=0.99\linewidth]{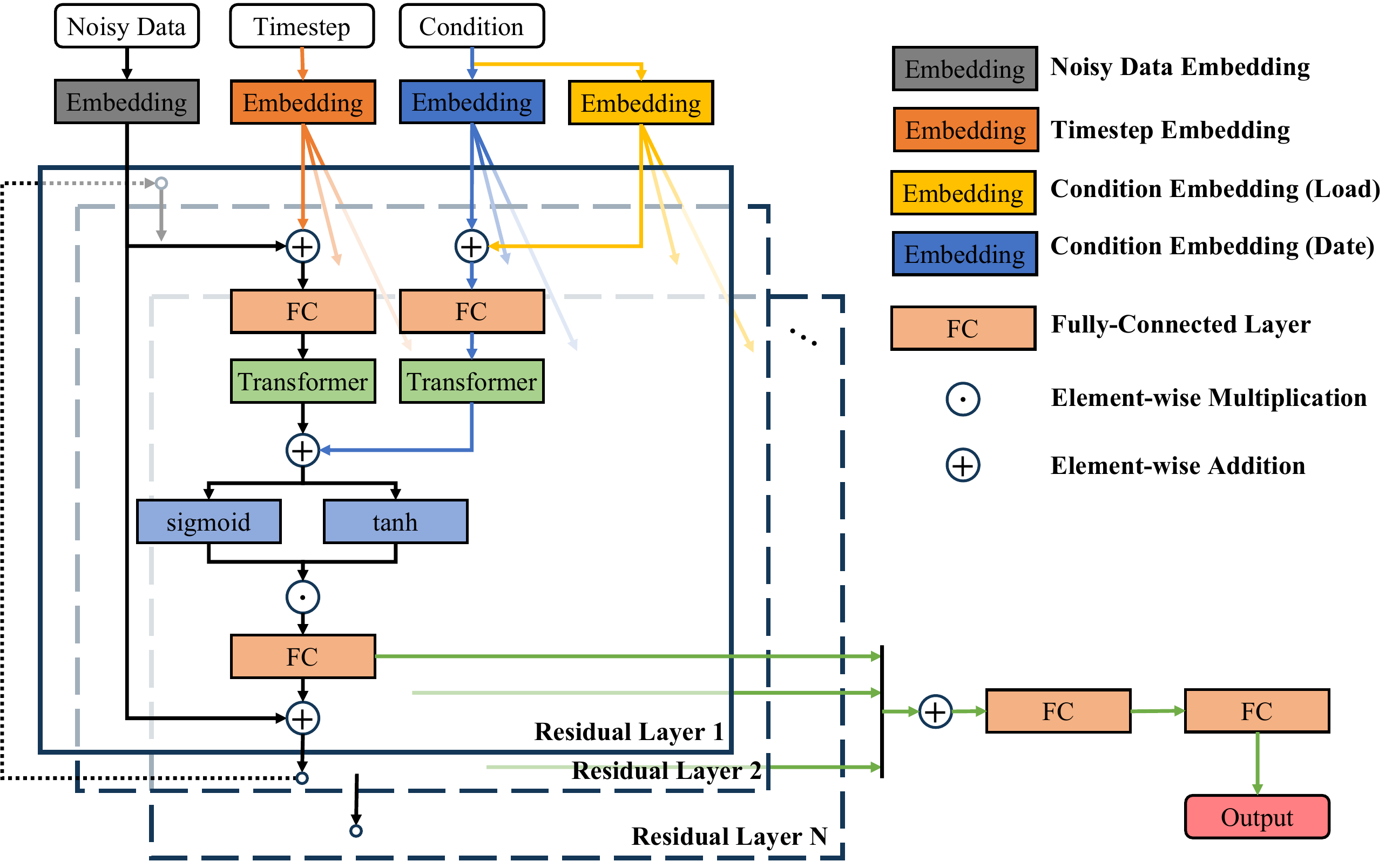}
    \caption{The neural network architecture of noise estimation model.}
    \label{fig_model}
\end{figure}

Because all inputs (e.g., noisy data and conditional information) of $\bm{\epsilon}_{\theta}$ in Eq. (\ref{loss_conditional}) are time-series, we design a deep learning model based on the attention mechanism\cite{vaswani2017attention} to predict the noises that are added to the original data during the diffusion process. The designed model is composed of N stacked residual layers, where the architecture is illustrated in Fig. \ref{fig_model}. In particular, the inputs include the latent variable $\bm{x}_t$, timestep $t$, and condition information $\bm{x}_c$. Furthermore, we aggregate the output of each residual layer to form the final model output (i.e., estimated noise) by using skip connections\cite{kong2021diffwave}. The designed model primarily consists of an embedding module, a residual module, and an output module, which are described respectively below.

\subsubsection{Embedding Module}
In order to preserve the temporal relationship of data, we perform the embedding transformation for each input\cite{vaswani2017attention}. First, we convert timestep $t$ into an encoding vector $t_{\text{emb}}$, because the model needs to yield different outputs $\bm{\epsilon}_{\theta}(\bm{x}_t, t \vert \bm{x}_{\text{c}})$ for different $t$:
\begin{equation}
    \begin{split}
        &\bm{t}_{\text{emb}} (2i) = \sin{ ( 10^{\frac{2i \text{x} 8}{d_{\text{model}}} t} ) }, \\
        &\bm{t}_{\text{emb}} (2i+1) = \cos{ ( 10^{\frac{2i \text{x} 8}{d_{\text{model}}} t} ) },
    \end{split}
\end{equation}
where $d_{\text{model}}$ denotes the encoding vector's dimensionality; and $i \in [0,\, d_{\text{model}}/2]$ is the index of data dimension.

Then, since $\bm{x}_t$ is from the same space as $\bm{x}_0$, it is also the sequence data. To get the corresponding embedding ${\bm{x}_t}_{\text{emb}}$, we apply the linear layer for data transformation:
\begin{equation}\label{sequence_data}
    {\bm{x}_t}_{\,\text{emb}} = \textit{FC}(\bm{x}_t) + \textit{PE}\big(\textit{FC}(\bm{x}_t)\big),
\end{equation}
where $\textit{FC}(\cdot)$ and $\textit{PE}(\cdot)$ denote the fully-connected layer and positional encoding function, respectively.

Finally, we consider that continuous conditional information $\bm{x}_c$ contains the customer's typical load ${\bm{x}_{c, \text{load}}}$ and target date ${\bm{x}_{c, \text{date}}}$, so we need to process them separately. For ${\bm{x}_{c, \text{load}}}$, we utilize the sequence data transformation in Eq. (\ref{sequence_data}), and then adopt the one-hot encoding method\cite{wang2024customer} for ${\bm{x}_{c, \text{date}}}$, as follows:
\begin{equation}
    \begin{split}
        &{\bm{x}_{c, \text{load}}}_{\text{emb}} = \textit{FC}(\bm{x}_{c, \text{load}}) + \textit{PE}\big(\textit{FC}(\bm{x}_{c, \text{load}})\big), \\
        &{\bm{x}_{c, \text{date}}}_{\text{emb}} = \textit{FC}\big(\textit{OneHot}(\bm{x}_{c, \text{date}})\big),
    \end{split}    
\end{equation}
where $\textit{OneHot}(\cdot)$ denotes the one-hot encoding function. In this manner, we improve the representation capability of all inputs and make them more suitable for model operation.

\subsubsection{Residual Module}
Since each embedding vector represents a different type of content, the naive addition or merging will reduce the information's diversity. Therefore, we deploy the attention mechanism to extract time-series characteristics of these vectors, and then combine them organically.

We divide the above four embedding vectors into two parts according to whether they are related to the condition, and then we fuse each part. Next, we distill temporal features of each part separately by using the attention function, followed by combining two parts to obtain the integrated feature $\bm{x}_{\text{total}}$:
\begin{equation}
    \begin{split}
        \bm{x}_{\text{total}} = &\textit{MHSA}\big(\textit{FC}({\bm{x}_t}_{\,\text{emb}} \oplus \bm{t}_{\text{emb}})\big)\, \oplus \\ 
        &\textit{MHSA}\big(\textit{FC}({\bm{x}_{c, \text{load}}}_{\text{emb}} \oplus {\bm{x}_{c, \text{date}}}_{\text{emb}})\big),
    \end{split}
\end{equation}
where operation $\oplus$ is the element-wise addition; and $\textit{MHSA}(\cdot)$ denotes the multi-head self-attention function. Henceforth, we utilize $\textit{FC}(\cdot)$ to enhance the non-linear fitting ability of the designed model, instead of representing data transformation.

To further facilitate the fusion of conditional information, we split $\bm{x}_{\text{total}}$ into two chunks, which will be processed by different activation functions, and then merge them again. We adopt different activation functions to avoid the information loss caused by the same processing operation, thus ensuring the diversity of information\cite{kong2021diffwave}. In addition, we also deploy the residual connection to effectively alleviate the vanishing gradient problem, as follows:
\begin{equation}
    \begin{split}
        \bm{x}_{\text{output}} &= \textit{FC}\big( \sigma(\bm{x}_{\text{sub},1}) \odot \tanh(\bm{x}_{\text{sub},2}) \big), \\
        \bm{x}_{\text{res}} &= \bm{x}_{\text{output}} \oplus {\bm{x}_t}_{\text{emb}},
    \end{split}
\end{equation}
where $\sigma(\cdot)$ and $\tanh(\cdot)$ denote the sigmoid and tanh activation function; $\bm{x}_{\text{sub},1}$ and $\bm{x}_{\text{sub},2}$ are two chunks of $\bm{x}_{\text{total}}$; and operation $\odot$ represents the element-wise multiplication.

For each residual layer, its $\bm{x}_{\text{res}}$ will serve as $\bm{x}_t$ of the next layer, and so on. Meanwhile, we collect $\bm{x}_{\text{output}}$ of each residual layer, which will work in the next module.

\subsubsection{Output Module}
As the designed model is composed of multiple stacked residual layers, some valuable information may be missed if we directly adopt the output of the last layer as the model result. Hence, we take account of each layer's output to jointly produce the ultimate output (i.e., the estimated noise), through integrating skip connections:
\begin{equation}
    \bm{\epsilon}_{\theta}(\bm{x}_t, t \vert \bm{x}_{\text{c}}) = \textit{FC} \big( \textit{Concat}(\bm{x}_{\text{skip},1}, \cdots ,\bm{x}_{\text{skip},N}) \big),
\end{equation}
where $\textit{Concat}(\cdot)$ is the vector concatenation operator; $\bm{x}_{\text{skip},i}$ denotes $\bm{x}_{\text{output}}$ of the $i$-th residual layer; and $N$ is the total number of residual layers.

\textbf{Remark 2.} \textit{The proposed model exploits the attention mechanism to extract features of all inputs separately and then combine them together, followed by using the skip connection to produce the estimated noise. The model comprehensively considers the distilled information from each estimation stage of the model so that it can realize accurate noise estimation.}

\subsection{Model Training and Synthesis Execution}
After establishing the neural network model for noise estimation, we train the designed model $\bm{\epsilon}_{\theta}$ by minimizing the loss function in Eq. (\ref{loss_conditional}). The detailed training process is summarized in Algorithm \ref{algo_training}. It should be mentioned that the noise schedule $\{\beta_{t}\}_{t=1}^{T}$ is a linear schedule based on the original implementation described in\cite{sohl2015deep_diffusion_models}, which is calculated by linear interpolation between 0.0001 and 0.5. In addition, to improve the quality and speed of training convergence, we use the Adam algorithm to optimize the model's parameters\cite{kingma2014adam_adam}, which adjusts the learning rate adaptively and uses momentum to amend the direction of parameter update.
\begin{algorithm}[!t]
    \caption{Conditional Diffusion Models Training}\label{algo_training}
    \begin{footnotesize}
    \Input{The model parameter $\theta$, diffusion total timestep $T$, noise schedules $\{\beta_{t}\}_{t=1}^{T}$, load profiles dataset $D$, condition dataset $D_c$, batch size $B$, total epoch number $E$, Adam parameters $\alpha^{\textit{Adam}}, \beta_{1}^{\textit{Adam}}, \beta_{2}^{\textit{Adam}}$.}
    \Output{The comprehensive trained model.}
    \textbf{Procedure:} \\
    \For{$e = 1, \dots, E$}{
        \For{each batch of training data}{
            \For{$i=1, \dots, B$}{
                Sample original data $\bm{x}_0 \sim D$, diffusion timestep $t \sim \text{Uniform}(1,T)$, Gaussian noise $\bm{\epsilon} \sim \mathcal{N}(\bm{0}, \bm{I})$; \\
                Select the corresponding condition $\bm{x}_{c}$ from $D_{c}$; \\
                Calculate $\bm{x}_t \leftarrow \sqrt{\overline{\alpha}_t} \bm{x}_0 + \sqrt{(1-\overline{\alpha}_t)} \bm{\epsilon}$; \\
                Obtain estimated noise $\bm{\epsilon}_{\theta}(\bm{x}_t, t \vert \bm{x}_{\text{c}})$; \\
                Compute model's loss $L^{(i)}\leftarrow \lVert (\bm{\epsilon} - \bm{\epsilon}_{\theta}(\bm{x}_t, t \vert \bm{x}_{\text{c}})) \rVert_{2}^{2}$; \\
            }
            Update model's parameters based on Adam algorithm $\theta \leftarrow \textit{Adam}(\nabla_{\theta} \frac{1}{B} \sum_{i=1}^{B} L^{(i)}, \alpha^{\textit{Adam}}, \beta_{1}^{\textit{Adam}}, \beta_{2}^{\textit{Adam}})$; \\
        }
    }
    \Return $\theta$
    \end{footnotesize}
\end{algorithm}

With comprehensive training, the proposed method can synthesize exclusive load profiles in an iterative manner according to customers' individual conditions, as presented in Algorithm \ref{algo_synthesis}. Specifically, we first determine the diffusion total timestep and accordingly obtain the noise schedules (see details in Section \ref{generation_method}). Then, we sample a random noise $\bm{x}_{T}$ from the Gaussian distribution. Moreover, we pick the customer's typical load and corresponding target date of data synthesis as conditional information $\bm{x}_{c}$, which will be used as the model's inputs with the random noise and timestep (see details in Section \ref{noise_model}). Subsequently, we calculate noisy data $\bm{x}_{t}$ based on the model's output, and repeat this process until $t=0$. At this moment, we take eventual denoised data $\bm{x}_0$ as the synthetic load profiles, which is customized for a customer.
\begin{algorithm}[!t]
    \caption{Customized Load Profiles Synthesis}\label{algo_synthesis}
    \begin{footnotesize}
    \Input{The well-trained model $\bm{\epsilon}_{\theta}$, diffusion total timestep $T$, noise schedules $\{\beta_{t}\}_{t=1}^{T}$, condition dataset $D_c$.}
    \Output{Customized synthetic load profiles.}
    \textbf{Procedure:} \\
        Sample $\bm{x}_T \sim \mathcal{N}(\bm{0}, \bm{I})$; \\
        Select typical load $\bm{x}_{c, \text{load}}$ and target date $\bm{x}_{c, \text{date}}$ according to the customer and its load profiles synthesis demand; \\
        Obtain $\bm{x}_c \leftarrow (\bm{x}_{c, \text{load}}\, ,\, \bm{x}_{c, \text{date}})$; \\
        \For{$t = T, \dots, 1$}{
            Sample $\bm{\epsilon} \sim \mathcal{N}(\bm{0}, \bm{I})$; \\
            \If{t = 0}{
                $\bm{\epsilon} \leftarrow \bm{0}$; \\
            }
            Calculate latent variable $\bm{x}_{t-1}$ according to Eqs. (\ref{bayes_posterior_conditional})--(\ref{parameter_conditional}) $\bm{x}_{t-1} \leftarrow \frac{1}{\sqrt{\alpha_t}} \Big(\bm{x}_t - \frac{\beta_t}{\sqrt{1-\overline{\alpha}_t}} \bm{\epsilon}_{\theta}(\bm{x}_t, t \vert \bm{x}_{\text{c}})\Big) + \frac{1-\overline{\alpha}_{t-1}}{1-\overline{\alpha}_t} \beta_t \bm{\epsilon}$; \\
        }
        \Return $\bm{x}_0$
    \end{footnotesize}
\end{algorithm}

\section{Case Studies}\label{sec_case}
In this section, we conduct extensive experiments to verify our proposed method. First, we introduce the experiment settings, including the dataset, evaluation metrics, benchmarks, and implementation details. Then, we demonstrate the performance of the proposed method by comparing it with four benchmarks in two scenarios, respectively. In addition, we conduct ablation experiments to validate the effectiveness of the proposed noise estimation model. Finally, we further evaluate the effectiveness of our proposed method for applying to the load forecasting task by comparing it with federated learning.

\subsection{Experiment Settings}
\subsubsection{Data Preparation}
We make use of the smart meter data collected in the Low Carbon London program for the numerical experiments \cite{schofield2015low_dataset}. This dataset contains electricity consumption data of 5198 customers from November 2011 to February 2014 with a time granularity of 30 minutes. After data preprocessing for missing values, we ultimately obtain full data across 1069 customers for the entire calendar year of 2013, i.e., 17520 pieces of recorded load data per customer. Since we consider daily load profiles in case studies, each sample contains 48 recorded load data. Moreover, all data are divided into the training, validation, and test sets, accounting for 60\%, 20\%, and 20\% of all data, respectively. Furthermore, we manually crawl the London weather data in 2013 for the load forecasting task.

\subsubsection{Evaluation Metrics}
Since it is challenging to measure the quality of synthetic data, especially for time series data \cite{huang2022dpwgan_GAN_privacy}, an effective evaluation indicator is necessary for load profiles synthesis. Moreover, the goal of the load profiles synthesis is to make the generated load data conform to the customer's load pattern, i.e., follow the distribution of real data. However, when the synthetic sample obeys the real data's distribution, the discrepancy between it and the real sample may still be large \cite{luo2018multivariate_difference}. Therefore, we design a comprehensive evaluation scheme that is suitable for both two scenarios. 

For the data generation scenario, we select four evaluation metrics from the perspective of sample level and distribution level, including root mean square error (RMSE), mean absolute error (MAE), maximum mean discrepancy (MMD), and Wasserstein distance (WD). The first two are sample-level classic metrics, and the latter two are used to quantify the difference between two distributions, expressed as follows:
\begin{equation}
    \text{MMD} = \frac{1}{M^2} \Big( \sum_{i=1}^{M} \sum_{j=1}^{M} \big( k(y_i, y_j) + k(\hat{y}_i, \hat{y}_j) - 2 \cdot k(y_i, \hat{y}_j) \big) \Big),
\end{equation}
\begin{equation}
    \text{WD} = \int_{-\infty}^{+\infty} \lvert F_1(z) - F_2(z) \lvert dz,
\end{equation}
where $y_i$ and $\hat{y}_i$ denote the $i$-th real and synthetic samples of data; $M$ is the total amount of data; $k(\cdot)$ denotes the Gaussian kernel function; $F_1(z)$ and $F_2(z)$ are the respective cumulative distribution functions of real and synthetic data. It is worth noting that a smaller value represents better data generation performance for these four metrics. In this way, the synthetic load profiles will be compared with real load profiles both numerically and distributionally to evaluate synthesis methods.

For the data augmentation scenario, there are generally two objectives: 1) minimizing the distribution shift from training data; and 2) increasing the amount of effective training data, thus improving the model generalization and performance. Considering that measuring the effect of augmented data is related to the downstream application task, we choose the load forecasting as the example in this paper. We adopt the \textit{Affinity} to quantify the impact of augmented data on the model decision boundary (i.e., performance), and the \textit{Diversity} to quantify the extent of augmented data on preventing the model overfitting, formulated as follows:
\begin{equation}
    \text{Affinity} = \mathcal{M}(g_{\omega}, D_{\text{val}}^{'}) - \mathcal{M}(g_{\omega}, D_{\text{val}}),
\end{equation}
\begin{equation}
    \text{Diversity} = \mathbb{E}_{D_{\text{train}}^{'}} L_{\text{train}},
\end{equation}
where $D_{\text{train}}$ and $D_{\text{val}}$ are the original training and validation datasets for the load forecasting task; $D_{\text{train}}^{'}$ and $D_{\text{val}}^{'}$ denote the corresponding augmented datasets; $g_{\omega}$ is the load forecasting model trained on $D_{\text{train}}$; $\mathcal{M}(g_{\omega}, D_{\text{val}})$ denotes the performance result of the trained data-driven model in the load prediction task evaluated on dataset $D_{\text{val}}$; and $L_{\text{train}}$ represents the model's training loss. Considering that augmented data will shift the decision boundary of data-driven models, which indicates that the value of affinity will not exceed 0, a larger value of affinity represents less impact on the model performance. In addition, a larger value of diversity implies a greater extent to which augmented data improve model training performance that can be improved by augmented data, and vice versa. To sum up, unlike before, the larger value of affinity and diversity stands for better augmentation performance.

\subsubsection{Benchmarks}
To verify the effectiveness and superiority of our proposed method, we select the following benchmarks based on the current state-of-the-art methods for comparison. Note that the first two are unconditional, while the latter two adopt the conditional data generation model.
\begin{itemize}
    \item Benchmark A: A GAN-based framework proposed in 2020 by Kababji et al.\cite{el2020data_GAN} for load pattern synthesis.
    \item Benchmark B: A least square GAN proposed in 2021 by Li et al.\cite{li2021privacy_LSGAN} for renewable scenario generation.
    \item Benchmark C: A ﬂow-based conditional generative model proposed in 2019 by Zhang et al. \cite{zhang2019scenario_CFGM} for scenario forecasting of residential load proﬁles.
    \item Benchmark D: A Wasserstein deep convolutional conditional GAN proposed in 2023 by Chen et al.\cite{chen2023federated_CGAN} to generate energy consumption data.
\end{itemize}

\subsubsection{Implementation Details}
The proposed method is implemented by the open source machine learning framework PyTorch\footnote{\url{https://pytorch.org}}. We conduct all experiments on an Ubuntu 18.04 LTS platform, which is equipped with the Intel Core i9-10980XE CPU and NVIDIA GeForce RTX 3090 GPU. Table \ref{implementation} summarizes the details of the model and training parameters.

\begin{table}[]
    \centering
    \caption{Implementation Details}
    \resizebox{0.99\linewidth}{!}{
    \begin{tabular}{c|c|c}
        \hline
        \textbf{Parameter} & \textbf{Definition} & \textbf{Value} \\
        \hline
        $T$ & diffusion total timestep & 50 \\
        ($\beta_1$, $\beta_T$) & noise schedule start and end & (0.0001, 0.5) \\
        $d_{\text{model}}$ & the embedding vector dimension & 16 \\
        $N$ & the number of residual layers & 6 \\
        $E$ & the number of training epochs & 200 \\
        $B$ & the batch size of training data & 16 \\
        $\alpha$ & the learning rate of Adam & 0.001 \\
        \hline
    \end{tabular}}
    \label{implementation}
\end{table}

\subsection{Performance in Data Generation Scenario}\label{DG}
In this part, we validate the performance of our proposed method in the data generation scenario by comparing it with the aforementioned four benchmarks. For the comprehensive validation, we randomly select 200 customers from the test dataset as cases, and hide their load data to imitate customers with little or no load data for the data generation scenario. After the load profiles synthesis, we release hidden load profiles and exploit the designed evaluation scheme to inspect the effect of synthetic load profiles from the sample and distribution levels. In addition, to avoid random errors, we repeat all experiments 10 times and then calculate the average values. The statistical results are summarized in Table \ref{result_generation}.
\begin{table*}[]
    \centering
    \caption{Numerical results of performance comparison under data generation scenario}
    \resizebox{0.9\linewidth}{!}{
    \begin{tabular}{c|ccccc}
        \hline
        \textbf{Metric} & \textbf{Benchmark A} & \textbf{Benchmark B} & \textbf{Benchmark C} & \textbf{Benchmark D} & \textbf{Proposed} \\
        \hline
        \textbf{RMSE (kW)} & 1.5584 (0.3099) & 1.5104 (0.2923) & 1.2271 (0.2153) & 1.1694 (0.2046) & \textbf{0.9806 (0.1512)} \\
        \textbf{MAE (kW)} & 1.2264 (0.2350) & 1.1571 (0.2138) & 0.9824 (0.1402) & 0.9205 (0.1304) & \textbf{0.7299 (0.1041)} \\
        \textbf{MMD} & 0.5727 (0.1023) & 0.5646 (0.09676) & 0.4986 (0.0904) & 0.4759 (0.0862) & \textbf{0.4092 (0.0667)} \\
        \textbf{WD} & 0.6526 (0.1366) & 0.6496 (0.1315) & 0.4968 (0.1080) & 0.4515 (0.0989) & \textbf{0.3582 (0.0749)} \\
        \hline
        \multicolumn{6}{l}{{\small*Each entry gives a pair of mean and standard deviation.}}
    \end{tabular}}
    \label{result_generation}
\end{table*}

It can be seen that our proposed method achieves the best performance in all metrics compared with benchmarks. The evaluation metric results of the proposed method are both less than 1 kW, and its MAE is even close to 0.7 kW. In contrast, the results of the four benchmarks are relatively poor, where the two error metrics (RMSE and MAE) are at least 19\% and 26\% higher, respectively. For the other two metrics (MMD and WD), the values of our proposed method are 0.41 and 0.36, whereas those of all benchmarks exceed 0.45. Moreover, the proposed method also has the smallest standard deviation in both sample-level and distribution-level metrics, with a maximum reduction of more than 50\% compared to the benchmarks. It is worth mentioning that both our proposed method and the latter two benchmarks achieve significant performance improvements compared to the first two unconditional benchmarks. This indicates the necessity of using conditional information for customized load profiles synthesis. Therefore, our proposed method attains the best load profiles synthesis performance in the data generation scenario in terms of accuracy and stability.

\begin{figure}
    \centering
    \subfigure[]{
        \centering
        \includegraphics[width=0.85\linewidth]{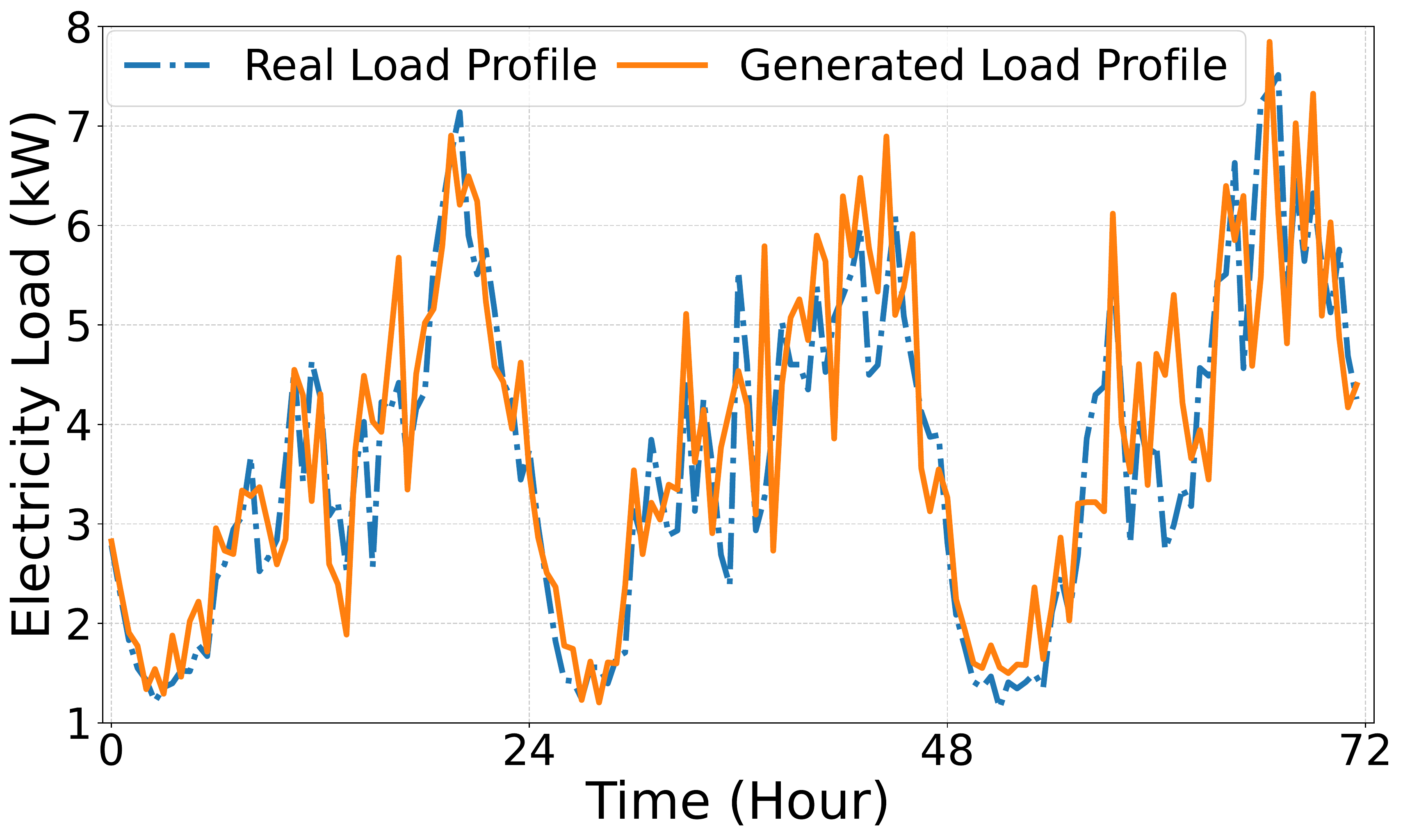}
        \label{fig_generation_one}
    }
    \subfigure[]{
        \centering
        \includegraphics[width=0.85\linewidth]{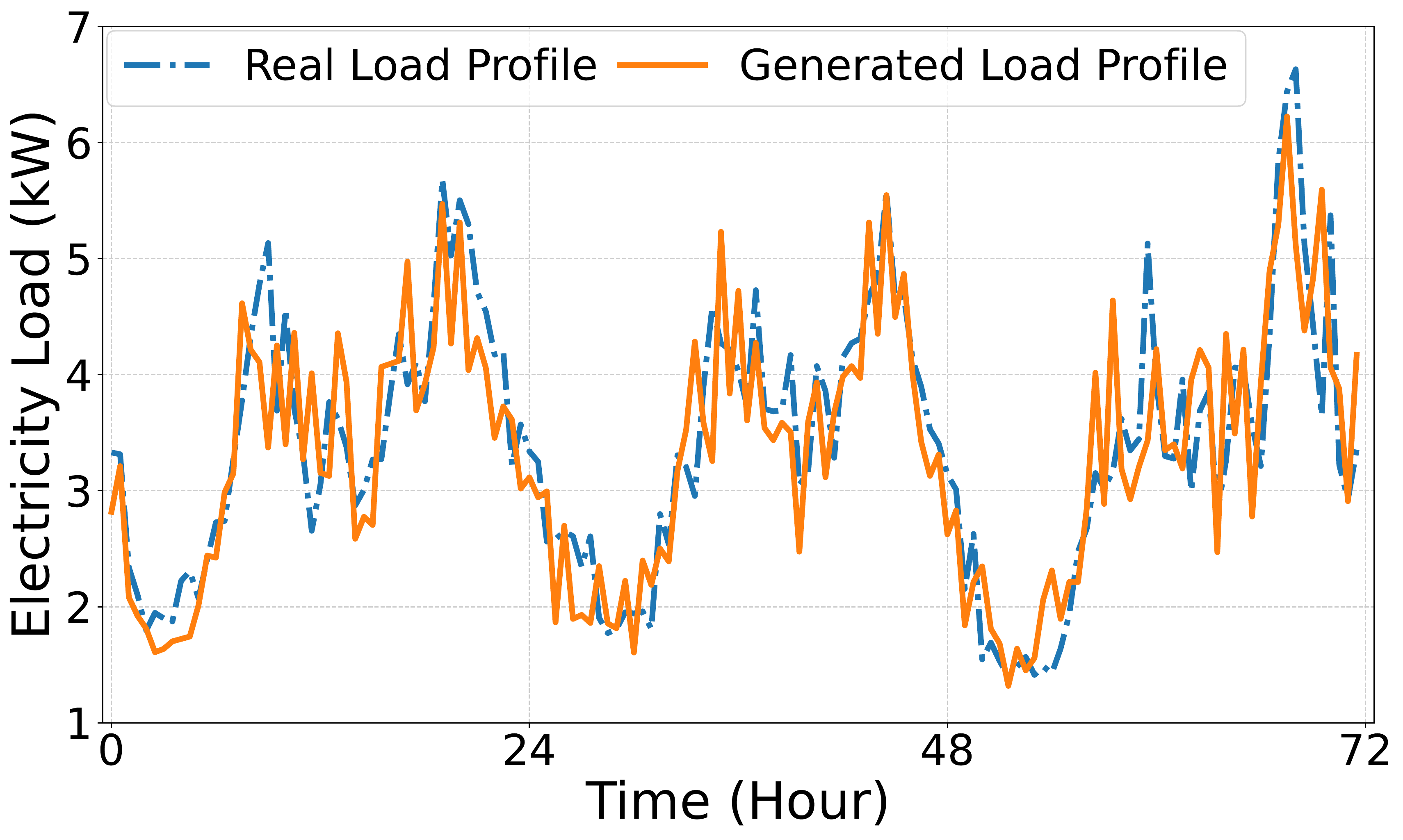}
        \label{fig_generation_two}
    }
    \caption{Load profiles synthesis results of two examples under data generation scenario in (a) summer and (b) winter.}
    \label{fig_generation}
\end{figure}

In order to provide a visual demonstration of the load profiles synthesis effects, we randomly choose one customer in summer and winter respectively as examples, and the visualization results are shown in Fig. \ref{fig_generation}. The generated load curves of the proposed method are close to the real curves for both the summer and winter cases, and they also follow the real load variation trends. Of particular importance is that our proposed method can accurately synthesize the daily peaks and valleys of load profiles, which is essential for the operation and control of power systems.

\begin{figure}
    \centering
    \includegraphics[width=0.9\linewidth]{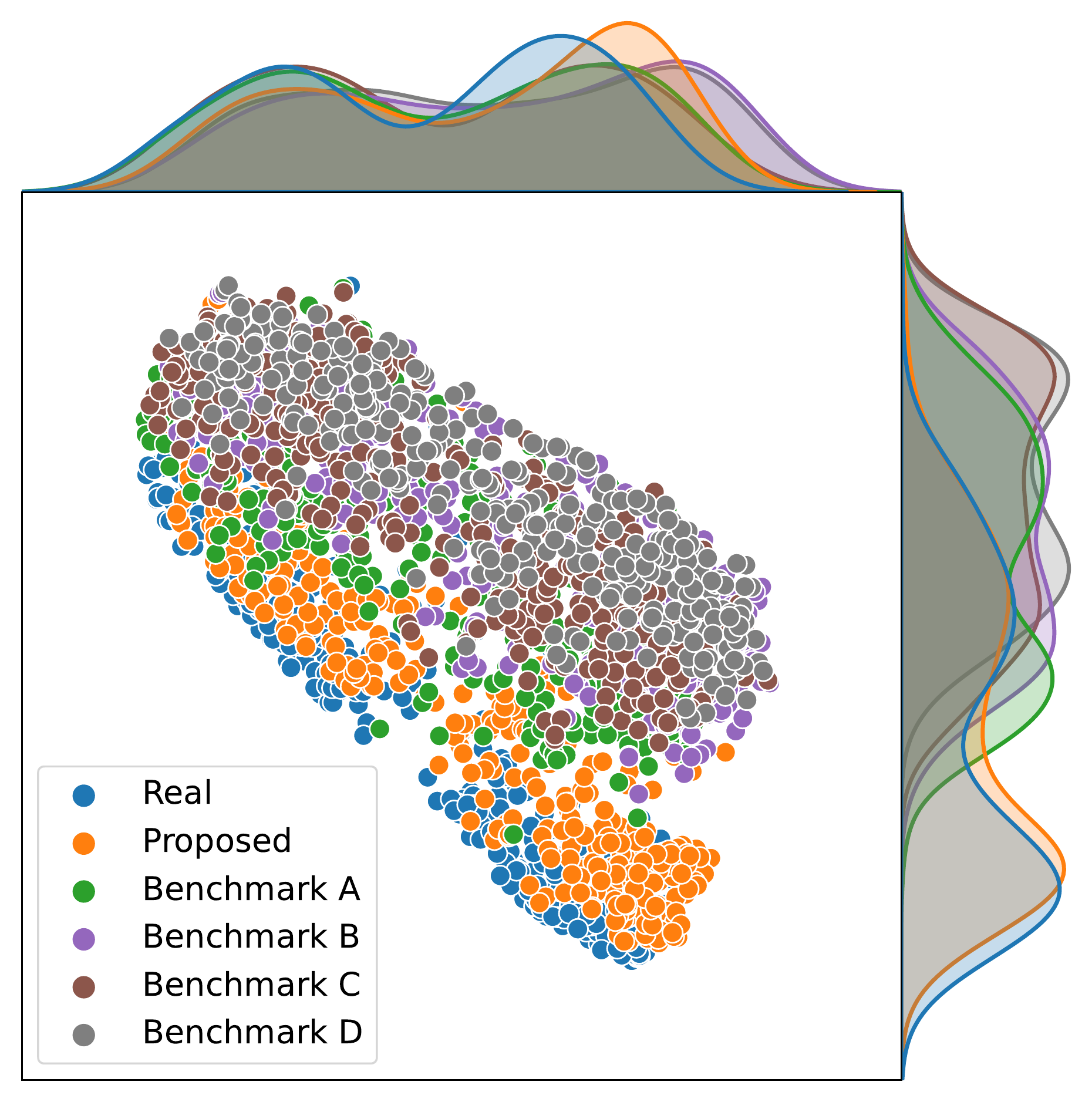}
    \caption{Two-dimensional visualization of synthesis performance comparison under data generation scenario. A data point represents one daily load profiles after dimensionality reduction. The curves on the axes depict the density curves for each type of load profiles in the two reduction dimensions.}
    \label{fig_generation_distribution}
\end{figure}
Similarly, to intuitively compare the performance of different methods, we randomly select another customer and make the 2-D visualization of its full-year synthetic load profiles that are generated by the proposed method and benchmarks, as shown in Fig. \ref{fig_generation_distribution}. This figure consists of a scatterplot in the center and density curves on the marginal axes. Specifically, each data point in the scatterplot represents a daily load profiles after dimensionality reduction to 2 dimensions by using umap\cite{mcinnes2018umap}. Each density curve on the axes describes the distribution of each type of scatter in two reduced dimensions. It is clear that the synthesis load profiles of our proposed method are closest to the real load profiles, and have the most similar shape to the real load profiles compared to other benchmarks. Moreover, from the perspective of density curves, the proposed method is also closest to the real load profiles in both dimensions, especially the vertical axis. This corresponds to the above four numerical metrics, thus demonstrating that our proposed method achieves the best synthesis performance.

\subsection{Performance in Data Augmentation Scenario}\label{DA}
\begin{figure}
    \centering
    \subfigure[]{
        \centering
        \includegraphics[width=0.85\linewidth]{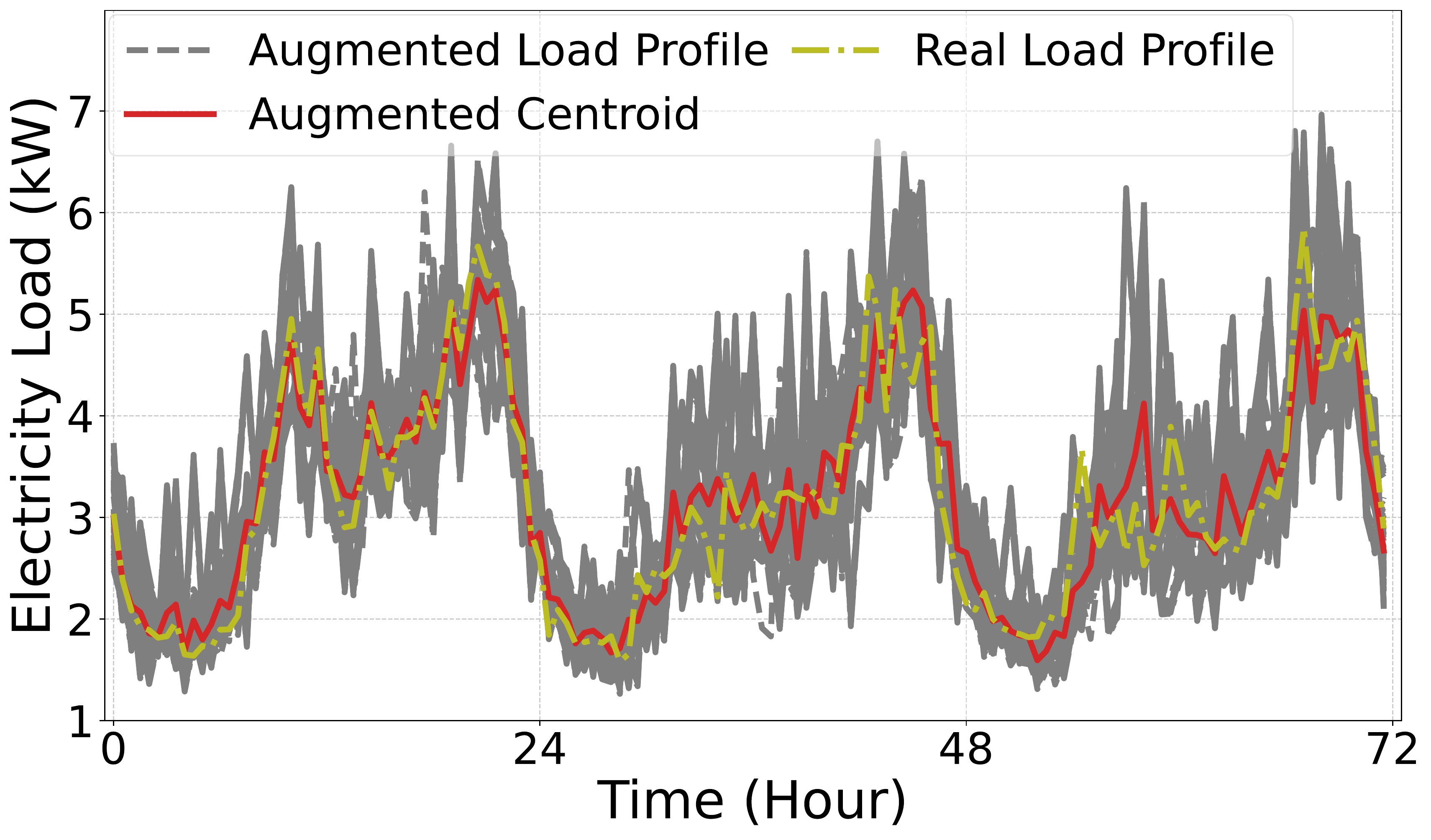}
        \label{fig_augmentation_one}
    }
    \subfigure[]{
        \centering
        \includegraphics[width=0.85\linewidth]{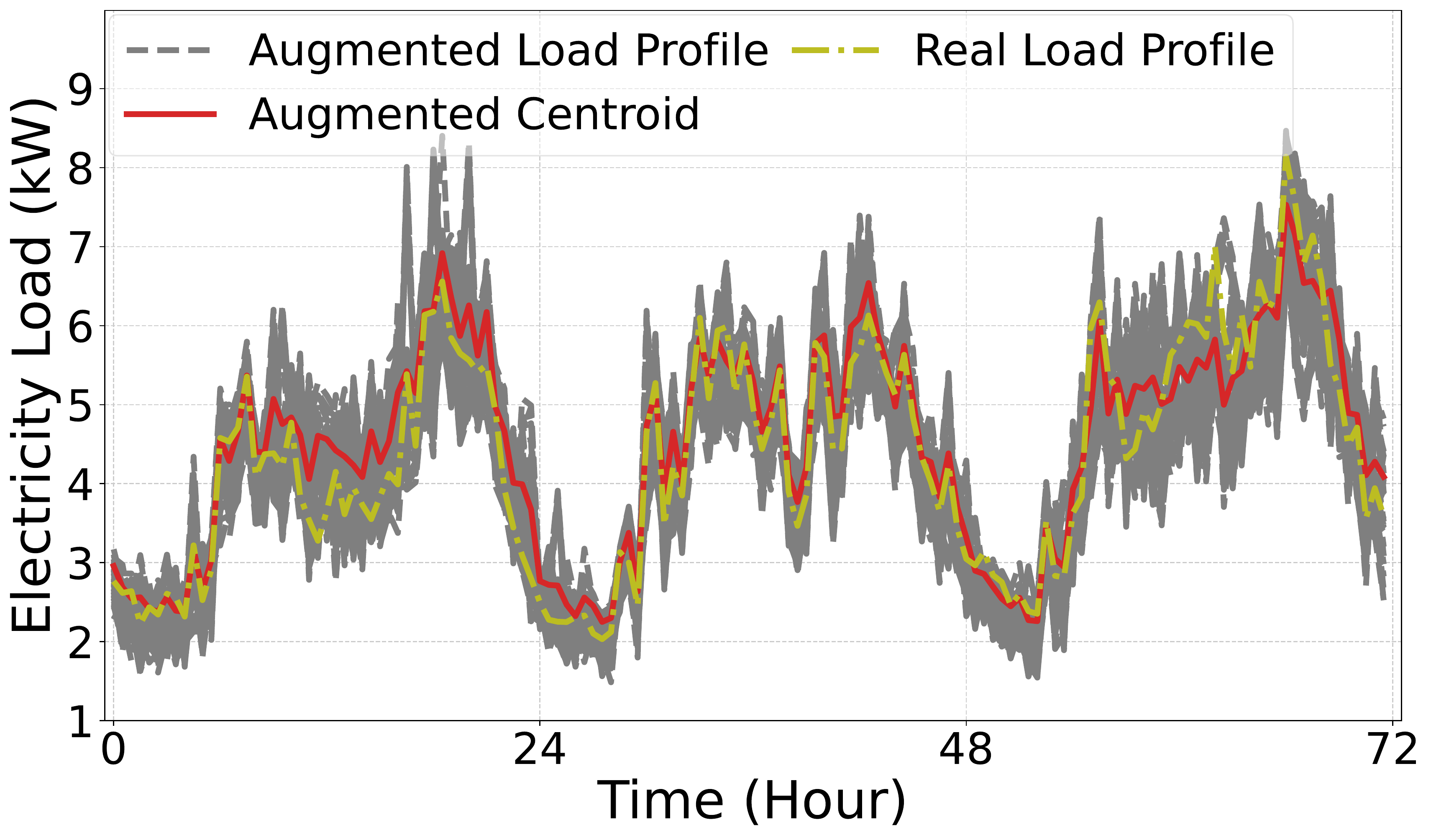}
        \label{fig_augmentation_two}
    }
    \caption{Load profiles synthesis results of two examples under data augmentation scenario in (a) summer and (b) winter.}
    \label{fig_augmentation}
\end{figure}

In this part, we further verify the performance of the proposed method in the data augmentation scenario. Following the same steps, we pick 200 customers without bias and augment each customer's full-year load profiles 50-fold. We present two augmentation examples using our proposed method, which are arbitrarily chosen and shown in Fig. \ref{fig_augmentation}. It should be noted that the gray curves represent augmented load profiles, and the gray area is formed by the superposition of 50 gray curves. Moreover, the augmented centroid is calculated by averaging 50 pieces of augmented load profiles. From the figure, the augmented load profiles not only cover the real load profile but also are mostly in the vicinity of the real curves, which proves the effectiveness of the data augmentation. Furthermore, the centroid of augmented data is also close to the real data, indicating that the statistical properties of the augmentation follow the ground truth.

\begin{figure}
    \centering
    \subfigure[]{
        \centering
        \includegraphics[width=0.95\linewidth]{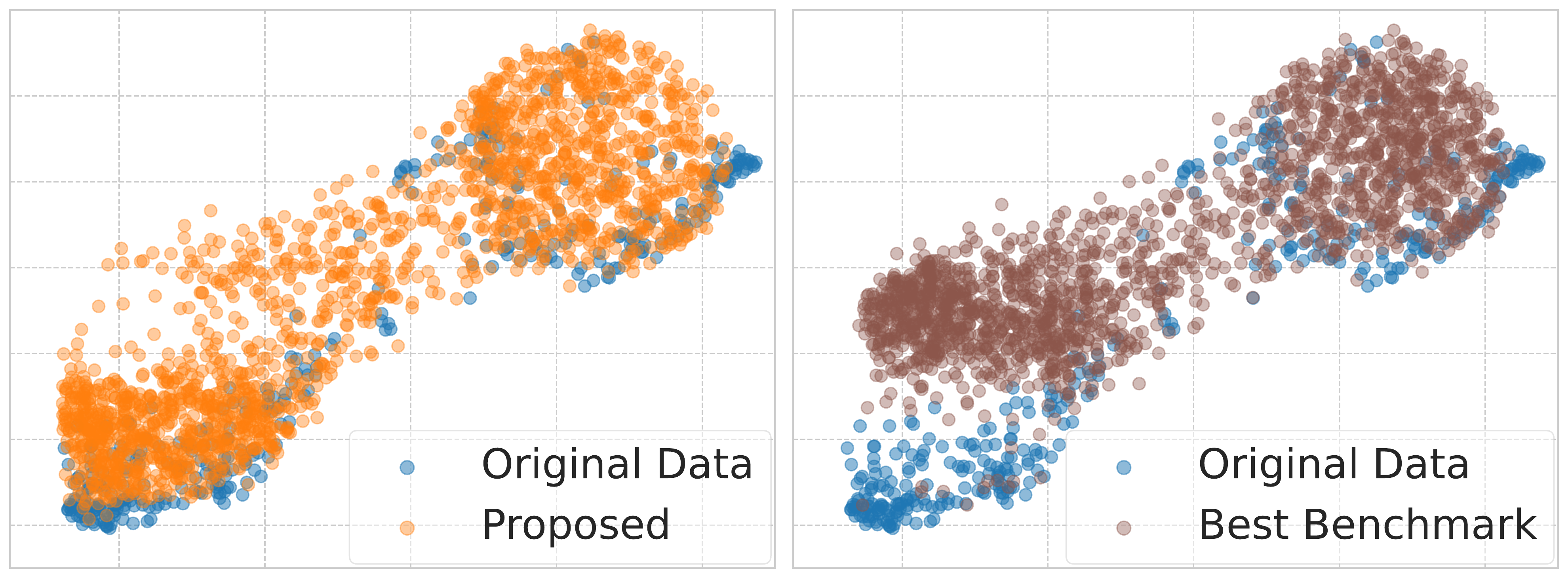}
        \label{fig_augmentation_one_dis}
    }
    \subfigure[]{
        \centering
        \includegraphics[width=0.95\linewidth]{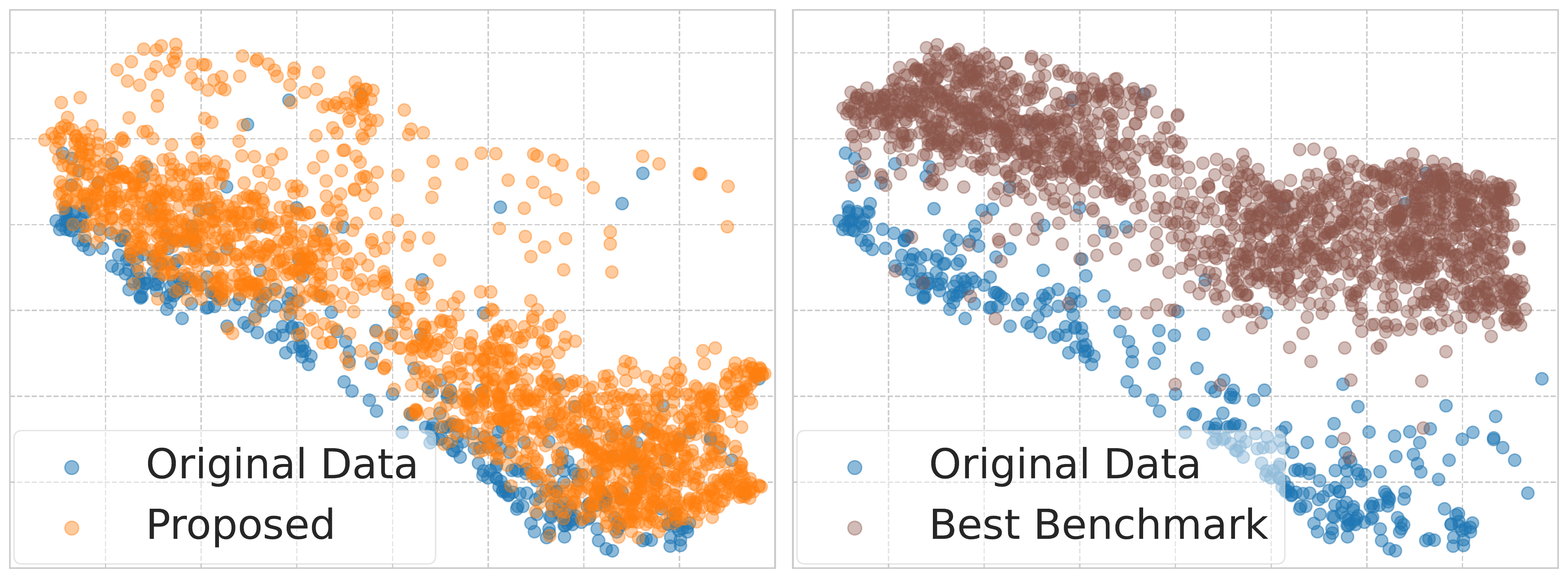}
        \label{fig_augmentation_two_dis}
    }
    \caption{Two-dimensional visualization of synthesis performance comparison with the best benchmark under data augmentation scenario. The original data are the measured real load profiles, and the other are augmented load profiles by using the proposed method or the best benchmark. A data point represents one daily load profiles after dimensionality reduction to two dimensions. The more similar shape and more overlap of the two types of data points signify the closer pattern of augmented load profiles and real load profiles.}
    \label{fig_augmentation_distribution}
\end{figure}

To explore the differences between distributions of the augmented and real load profiles, we employ umap for dimension reduction once again, and the visualization results of two examples are presented in Fig. \ref{fig_augmentation_distribution}. For the sake of clarity, we only select the best benchmark in experiments for comparison. Compared with the best benchmark, the augmented data of our proposed method are more similar in shape to the original data and have more sample point intersections. However, these two types of data are not entirely overlapping, thus the proposed method does not purely simulate original data, but learns the intrinsic manifold of load profiles, making it capable of generating augmented data with diversity.

\begin{table}[]
    \centering
    \caption{Performance comparison under data augmentation scenario}
    \resizebox{0.99\linewidth}{!}{
    \begin{tabular}{c|ccc}
        \hline
        \textbf{Method} & \textbf{Affinity} & \textbf{Diversity} & \textbf{Improvement$^1$ (\%)}\\
        \hline
        \textbf{Benchmark A} & -0.2922 & 0.7997 & 3.9882 \\
        \textbf{Benchmark B} & -0.2340 & 0.8634 & 4.1311 \\
        \textbf{Benchmark C} & -0.1293 & 1.3521 & 7.3723 \\
        \textbf{Benchmark D} & -0.1108 & 1.4563 & 7.5538 \\
        \textbf{Proposed} & \textbf{-0.0775} & \textbf{1.6381} & \textbf{8.9184} \\
        \hline
        \multicolumn{4}{l}{\textbf{\small*Each entry gives the mean value.}} \\
        \multicolumn{4}{l}{\textbf{\small$^1$This represents performance improvement in load forecasting.}}
    \end{tabular}}
    \label{result_augmentation}
\end{table}

To reveal the role of load profiles synthesis on real-world data analytics applications, we train the respective models for the load forecasting task, using the augmented data from each benchmark and our proposed method, respectively. Then, we exploit the designed evaluation metrics (i.e., affinity and diversity) to quantify the data augmentation performance. Furthermore, we train the load forecasting baseline model with original real load profiles, and accordingly calculate the performance improvement of load forecasting models after using the augmented data. All numerical results are listed in Table \ref{result_augmentation}. It can be observed that the proposed method has the largest affinity value exceeding $-0.1$, as well as the highest diversity value with at least 10\% improvement over the benchmarks. Therefore, our proposed method can complement effective training data that have minimal impact on the model decision boundary and improve model generalization. Therefore, our proposed method can complement effective training data that have minimal impact on the model performance and improve model generalization best. In addition, the performance improvement results further prove that our proposed method can enhance the model performance on the load forecasting task, with an extent of nearly 9\% on prediction accuracy.

\subsection{Performance Comparison for Ablation Experiments}
To further demonstrate the effectiveness of the proposed noise estimation model, we conduct ablation experiments to reveal the impact of key components (i.e., the attention mechanism and skip connections) on model performance. According to the structure of our proposed model, we construct three new benchmarks, as follows:
\begin{itemize}
    \item Benchmark E1: the Transformer in the residual module is substituted by the long short-term memory (LSTM) network, and the skip connections operation in the output module is deleted.
    \item Benchmark E2: the Transformer in the residual module is substituted by the LSTM network, but the skip connections operation in the output module is reserved.
    \item Benchmark E3: the Transformer in the residual module is reserved, but the skip connections operation in the output module is deleted.
\end{itemize}

Note that when the skip connections operation is removed, the output of benchmarks is the output of the last residual layer in their models. In addition, our proposed method and benchmarks utilize the same data for performance comparison.
\begin{table}[]
    \centering
    \caption{Performance comparison for Ablation Experiment}
    \resizebox{0.99\linewidth}{!}{
    \begin{tabular}{c|cccc}
        \hline
        \textbf{Metric} & \textbf{E1} & \textbf{E2}& \textbf{E3} & \textbf{Proposed}\\
        \hline
        \textbf{RMSE} & 1.7018 & 1.3509 & 1.6165 & \textbf{0.9806} \\
        \textbf{MAE} & 1.3460 & 1.0262 & 1.3356 & \textbf{0.7299} \\
        \textbf{MMD} & 0.6686 & 0.5569 & 0.6354 & \textbf{0.4092} \\
        \textbf{WD} & 0.7809 & 0.5921 & 0.7212 & \textbf{0.3582} \\
        \hline
        \textbf{Affinity} & -0.3965 & -0.1875 & -0.3273 & \textbf{-0.0775} \\
        \textbf{Diversity} & 0.5016 & 0.9863 & 0.6185 & \textbf{1.6381} \\
        \hline
        \multicolumn{5}{l}{\textbf{\small*Each entry gives the mean value.}} \\
        \multicolumn{5}{l}{\textbf{\small**RMSE and MAE are all in kilowatts(kW).}}
    \end{tabular}}
    \label{ablation_experiment}
\end{table}

We repeat the experiments 10 times to calculate the average values, and the results are shown in Table \ref{ablation_experiment}. It can be seen that the performance degradation is significant, with the absence of the attention mechanism or skip connections. Specifically, compared to our proposed method, when we remove only the attention mechanism or skip connections, the RMSE of benchmarks increases by 37.8\% and 64.8\%, respectively. Similarly, their diversity metrics drop by 39.8\% and 62.2\%, respectively. This demonstrates the importance of the attention mechanism and skip connections on the model performance, and the attention mechanism is more crucial than skip connections. In addition, when we remove both the attention mechanism and skip connections, the synthesis performance further declines. For example, compared to Benchmark E2 and E3, the MMD of Benchmark E1 raises by 20.1\% and 5.2\%, respectively. Likewise, its affinity metric also decreases by 111.5\% 21.1\%, respectively. This further indicates that the attention mechanism has more contribution to the model performance. Therefore, even with the same structure, the model performance will suffer an obvious degradation without the attention mechanism and skip connections.

\subsection{Performance Comparison with Federated Learning in Load Forecasting Task}
In this part, we evaluate the performance of our proposed method on downstream application tasks, where the load forecasting task is selected as an example. We randomly select 50 customers with a full year of load profiles for the experiment. Considering that the historical load data of a single customer is insufficient for training a data-driven model, we adopt the load profiles synthesis and federated learning\cite{wang2022privacy_self} to deal with the data shortage problem, respectively. Specifically, on the one hand, we utilize our proposed method and four benchmarks to augment each customer's load profiles 50-fold respectively, and then train each customer's individual load forecasting model by using augmented data. On the other hand, we adopt federated learning to collaboratively train a global load forecasting model among the 50 customers. In this way, both two approaches increase the amount of training data in essence. It is worth noting that for a single customer, the amount of training data for the load forecasting model is the same in the above two approaches. Furthermore, we use RMSE as the metric to evaluate the performance of trained load forecasting models in the load forecasting task.

\begin{figure}
    \centering
    \includegraphics[width=0.99\linewidth]{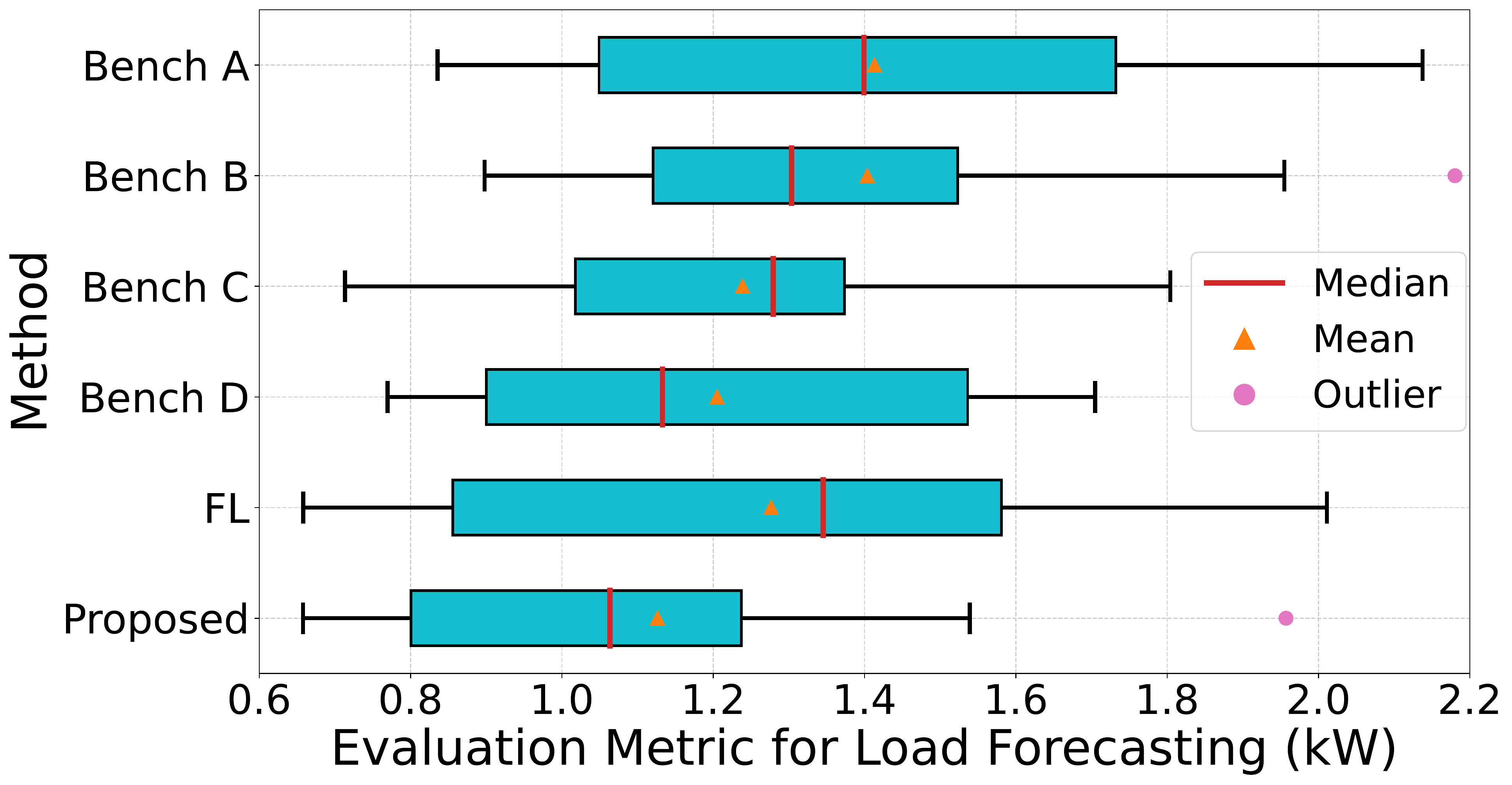}
    \caption{Performance comparison of the proposed method, benchmarks, and federated learning in the load forecasting task. RMSE is selected as the performance evaluation metric for load forecasting. FL stands for federated learning, and Bench is short for ``Benchmark".}
    \label{fig_comparison_FL}
\end{figure}

We summarize the load forecasting results of the 50 customers and calculate the statistical values, which are shown in Fig. \ref{fig_comparison_FL}. It is clear that our proposed method outperforms the benchmarks in load forecasting. This makes sense since the forecasting performance is determined by the quality of synthesized load profiles (see details in Sections \ref{DG} and \ref{DA}). The model trained using augmented data from our proposed method achieves the lowest mean and median of RMSE, implying the highest forecasting accuracy. Moreover, the RMSE range of the federated learning is wider than the proposed method, indicating that the global model's accuracy for each customer is relatively poor. This is because the global model trained through federated learning loses some accuracy to guarantee generalization. Therefore, at the customer level, our proposed method is a better choice to improve the model performance on real-world data analytics applications.

\section{Conclusion}\label{sec_conclusion}
In this paper, we concentrate on customized load profiles synthesis for electricity customers. High-quality customized synthetic load profiles enable customers to build personalized data-driven models for various data analytics applications, e.g., load forecasting and energy management. We propose a novel customized load profiles synthesis method based on conditional diffusion models, which generates tailored load profiles conditioned on the customer's load characteristics and individual demands. Furthermore, we design a noise estimation model with the attention mechanism and skip connection to implement the conditional diffusion models. 
Case studies comprehensively verify that our proposed method can realize customized load profiles synthesis and outperform existing methods. The average enhancement of evaluation metrics is 17.5\% and 21\% in the data generation and data augmentation scenarios, respectively. Moreover, the proposed method aids customers in improving the performance of their data-driven models on application tasks, such as load forecasting, by providing synthetic training data.

Due to the widespread deployment of smart meters and the increasing attention on data security, how to design efficient and secure data protection approaches is an important research direction. We will consider this topic in our future work.

\bibliographystyle{ieeetr}
\bibliography{ref}

\begin{IEEEbiography}[{\includegraphics[width=1in,height=1.25in,clip,keepaspectratio]{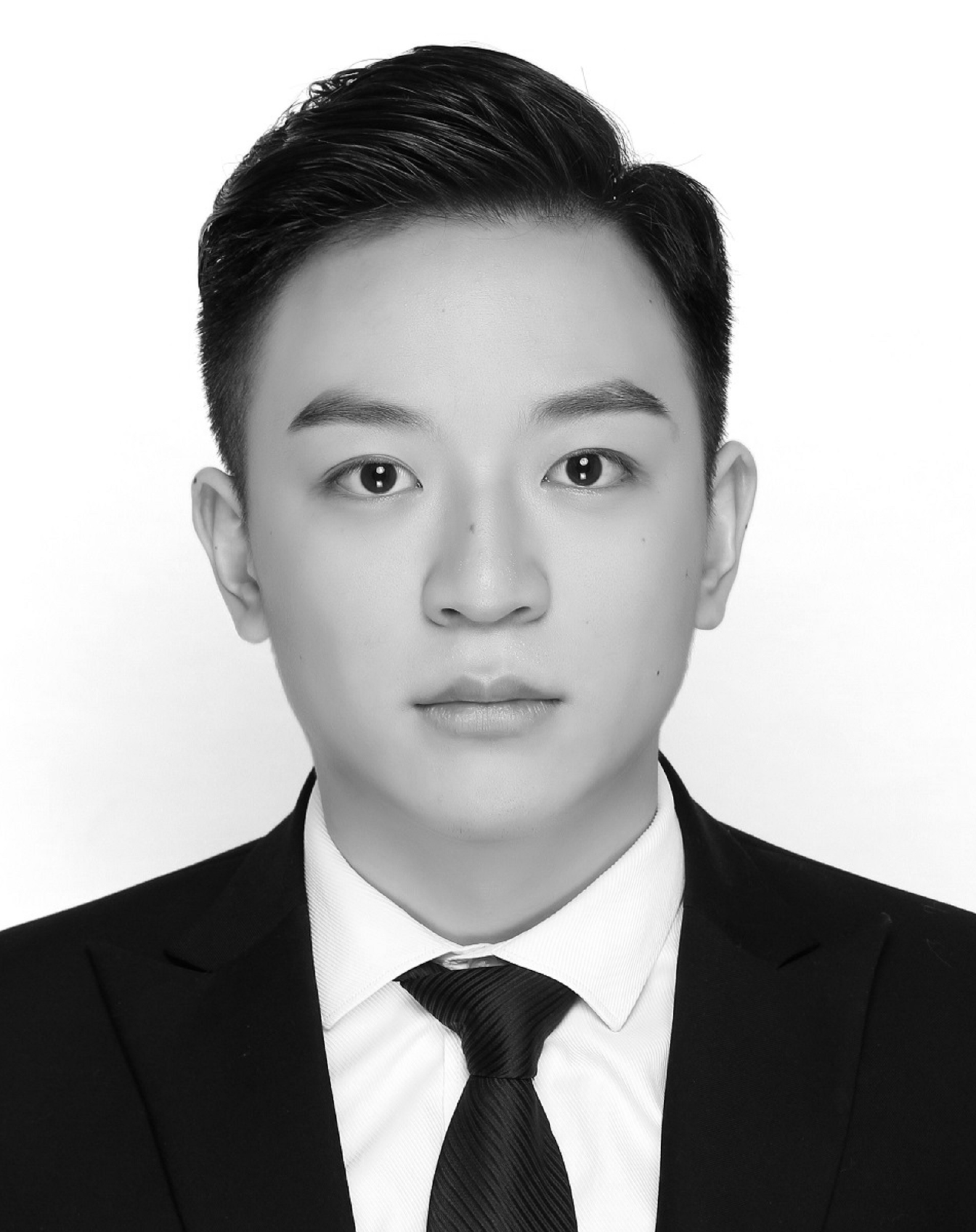}}]{Zhenyi Wang}(S'22)
	received the B.E. degree in cybersecurity from Sichuan University, Chengdu, China, in 2021.
	He is currently pursuing the Ph.D. degree in electrical and computer engineering at University of Macau, Macao, China.
 
 	His research interests include data analytics in demand response and electricity market, trustworthy and data-centric AI for power systems, and the intersection of causal inference with AI.
\end{IEEEbiography}
\begin{IEEEbiography}[{\includegraphics[width=1in,height=1.25in,clip,keepaspectratio]{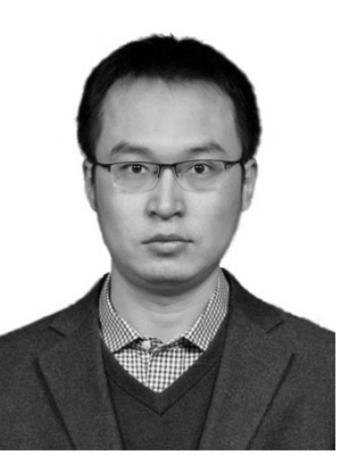}}]{Hongcai Zhang}(S'14--M'18--SM'23)
	received the B.S. and Ph.D. degree in electrical engineering from Tsinghua University, Beijing, China, in 2013 and 2018, respectively. 
	He is currently an Assistant Professor with the State Key Laboratory of Internet of Things for Smart City and Department of Electrical and Computer Engineering, University of Macau, Macao, China. In 2018-2019, he was a postdoctoral scholar with the University of California, Berkeley. He is an Associate Editor of IEEE Transactions on Power Systems and Associate Editor of Journal of Modern Power Systems and Clean Energy. 
	
	His current research interests include Internet of Things for smart energy, optimal operation and optimization of power and transportation systems, and grid integration of distributed energy resources. 
\end{IEEEbiography}

\end{document}